\documentclass[11pt,letterpaper]{article}
\usepackage[margin=1in]{geometry}
\usepackage{microtype}
\usepackage{amsmath,amssymb,amsthm,mathtools}
\usepackage{booktabs,graphicx}
\usepackage{placeins}
\usepackage{tikz}
\usetikzlibrary{arrows.meta}
\definecolor{figblue}{HTML}{245774}
\definecolor{figlight}{HTML}{EAF1F5}
\usepackage[round]{natbib}
\usepackage[hidelinks]{hyperref}
\hypersetup{
  pdftitle={Tight Stochastic Sequential Covers for General VC Classes with Unknown i.i.d. Features},
  pdfauthor={Changlong Wu and Wojciech Szpankowski}
}

\newcommand{\Hc}{\mathcal H}
\newcommand{\Fc}{\mathcal F}
\newcommand{\Gc}{\mathcal G}
\newcommand{\Xc}{\mathcal X}
\newcommand{\Qc}{\mathcal Q}
\newcommand{\Pc}{\mathcal P}
\newcommand{\E}{\mathbb E}
\newcommand{\Prb}{\mathbb P}
\newcommand{\one}{\mathbf 1}
\newcommand{\vc}{\operatorname{VC}}
\newcommand{\pdim}{\operatorname{Pdim}}
\newcommand{\pos}[1]{\left(#1\right)_{+}}
\newcommand{\harm}{\mathsf H}
\newcommand{\R}{\mathbb R}

\newtheorem{theorem}{Theorem}
\newtheorem{lemma}[theorem]{Lemma}
\newtheorem{proposition}[theorem]{Proposition}

\theoremstyle{definition}
\newtheorem{definition}[theorem]{Definition}

\newtheorem{example}[theorem]{Example}

\title{Tight Stochastic Sequential Covers for General VC Classes with Unknown i.i.d. Features}
\author{
  Changlong Wu\textsuperscript{1}\qquad
  Wojciech Szpankowski\textsuperscript{2,3}\\[0.6em]
  {\normalsize\textsuperscript{1}\,University of Arizona}\\
  {\normalsize\textsuperscript{2}\,Purdue University}\\
  {\normalsize\textsuperscript{3}\,Jagiellonian University}\\[0.4em]
  {\small\textsuperscript{1}\,\href{mailto:clwu@arizona.edu}{\texttt{clwu@arizona.edu}}}\\
  {\small\textsuperscript{2,3}\,\href{mailto:szpan@purdue.edu}{\texttt{szpan@purdue.edu}}}
}
\date{}

\begin{document}
\maketitle

\begin{abstract}
We study hybrid online learning with stochastic features and adversarial labels when the feature distribution is unknown. Prior work gave stochastic sequential covers of log-size $O(d\log^2 T)$ for a general VC class and logarithmic bounds only under stronger star-type structure. We show that finite VC dimension alone is enough: every binary class of VC dimension $d$ admits a distribution-blind exact cover with log-size $O(d\log T+\log(1/\delta))$, matching the worst-case $\Omega(d\log(T/d))$ dependence on the horizon. The key ingredient is a new local cube weight that counts Boolean subcubes incident to each realizable trace. Unlike the equal splits used in earlier constructions, we split mass unevenly according to these cube weights; a coordinate-deletion inequality shows that VC dimension controls the total cost of these splits. This yields, to our knowledge, the first tight $O((d+1)\log T)$ expected worst-case log-loss regret for general VC partitions with unknown i.i.d. features, and an exact polynomial-time implementation for planar affine halfspaces. We also show that the binary phenomenon \emph{does not} extend from VC dimension to pseudo-dimension: a fixed pseudo-dimension-one sigmoid class requires $\Omega(\log^2 T)$ bits for uniform sequential approximation at accuracy $1/T$.
\end{abstract}

\section{Introduction}

Hybrid online learning studies sequential prediction when the features are stochastic but the labels may be adversarial \citep{lazaric2012}. We focus on the distribution-blind setting: the feature distribution is unknown, features arrive only sequentially, and a single prediction strategy must work for every i.i.d. feature law. This differs from two commonly studied settings: known-distribution learning, where the feature law is available to the learner, and transductive prediction, where the full feature design is revealed in advance \citep{bhatt2023smoothed,qian2026}.

A central tool in this setting is the stochastic global sequential cover introduced by \citet{wu2023}. Such a cover is a finite collection of sequential predictors that, with high probability over the feature sequence, simultaneously reproduces the trace of every hypothesis in the class. For a class of VC dimension $d$, the general bound of \citet{wu2023} is
$\log|\Gc|=O(d\log^2T)$. They obtained the sharper $O(\log T)$ dependence only under additional structure, first through finite star number and then through Star--Littlestone dimension and special compositions such as intervals and cuboids. These conditions are genuinely restrictive: none of the arguments covered even affine halfspaces in $\R^2$, which were singled out as an open case in \citet[Remark~4]{wu2023}. Thus it was unclear whether the extra logarithm was an artifact of the proof or a real price for working with a general VC class.

Our main result shows that the extra logarithmic factor is only an artifact. Every binary class of finite VC dimension admits a polynomial stochastic sequential cover. More precisely, for every $\Hc$ with $\vc(\Hc)=d$,
\begin{equation}\label{eq:intro-main}
\log|\Gc|
\leq
6d\harm_T+\log\Pi_{\Hc}(T)
+\log(1/\delta)+\log3,
\end{equation}
where $\harm_T=\sum_{t=1}^T 1/t$ and $\Pi_{\Hc}(T)$ is the growth function. By Sauer--Shelah, this is $O(d\log T+\log(1/\delta))$. We further exhibit a sparse-set class that requires cover log-cardinality $\Omega(d\log(T/d))$. Thus, the logarithmic dependence on $T$ is tight in the worst case.

Our central conceptual contribution is a new combinatorial quantity that controls how probability mass should be split at a branch. Earlier logarithmic-cover constructions~\citep{wu2023} build a realization tree and, whenever both labels remain feasible, split the available covering functions equally between the two children. Each such split costs one bit, so obtaining an \(O(\log T)\) cover requires a structural argument showing that only \(O(\log T)\) relevant splits can occur; this is precisely where star-type assumptions enter. We keep the same basic splitting viewpoint but abandon equal splitting. Each realizable trace is assigned a local weight given by the number of Boolean subcubes incident to it, and the two children receive mass in proportion to these weights. A new coordinate-deletion inequality shows that VC dimension alone controls the total cost of these unequal splits. Under a random ordering, the expected cost is \(O(d/t)\) when \(t\) coordinates remain, which sums to \(O(d\log T)\). This provides the key argument that extends the result from the previously star-type classes to every finite-VC class.

The covering bound \eqref{eq:intro-main} has two immediate consequences. First, it essentially resolves the open problem posed by \citet{bhatt2021} for the two-level class \(\Fc_{\Hc}=\{x\mapsto a+(b-a)h(x):h\in\Hc,\ a,b\in[0,1]\}\). For an unknown feature distribution, \citet{bhatt2021} established a general VC-class regret bound of square-root order. Using the stochastic sequential covers, \citet{wu2023} subsequently improved this to \(O((d+1)\log^2 T)\). Our result further sharpens the rate to \(O((d+1)\log T)\), which has the optimal logarithmic dependence on \(T\) for worst-case VC classes. To our knowledge, an \(O(d\log T)\) guarantee was not previously known even under the weaker i.i.d.-label setting; for example, the general VC-class bounds of \citet{bilodeau2023density} still lose logarithmic factors. Here, our tight logarithmic rate holds in the stronger setting where the feature distribution is unknown and the labels may be chosen adversarially after observing the entire feature sequence.

Second, the improvement changes the computational picture. The earlier
general bound~\citep{wu2023} gives a cover of size
$\exp(O(d\log^2 T))=T^{O(d\log T)}$, which is superpolynomial in $T$
even when the dimension is fixed; in particular, naive aggregation over
that cover is already superpolynomial for planar linear separators. Our
bound reduces the cover size to $T^{O(d)}$. Moreover, for affine
halfspaces in $\R^2$, we give a more direct exact polynomial-time
implementation that avoids explicitly enumerating the cover.

Finally, we show that the result \emph{does not} extend to arbitrary
real-valued classes by simply replacing VC dimension with
pseudo-dimension. We construct a fixed sigmoid family of
pseudo-dimension one for which every distribution-blind uniform
sequential $1/T$-cover has log-cardinality $\Omega(\log^2 T)$. Thus,
the improvement from $\log^2 T$ to $\log T$ is a genuinely binary
phenomenon rather than a generic consequence of finite dimension.

\subsection{Related work}\label{sec:related}

Table~\ref{tab:protocols} summarizes the closest regret guarantees. \citet{bhatt2021} study the same two-level comparator under conditionally Bernoulli labels, while \citet[Theorem 11]{bhatt2023portfolio} allow adversarial outcomes but retain a square-root term for general VC classes. Logarithmic rates are known under stronger information assumptions, such as a known feature distribution or advance access to the full feature design \citep{bhatt2023smoothed,qian2026}. The difficulty of removing distributional knowledge also appears in smoothed online learning: most earlier results assume that the reference measure controlling the adversary is known, whereas \citet{blanchard2025agnostic} develop a new recursive-cover method to obtain agnostic sublinear regret when that reference measure is unknown. Our result establishes the logarithmic rate for general finite-VC partition classes while retaining both an unknown feature distribution and sequential feature access.

\begin{table}[t]
\centering
\footnotesize
\setlength{\tabcolsep}{3pt}
\begin{tabular}{@{}llll@{}}
\toprule
Work & Feature law & Access & Rate \\
\midrule
\citet{bhatt2021} & unknown & causal & $\sqrt{dT\log T}$ \\
\citet{bhatt2023portfolio} & unknown & causal & square-root \\
\citet{bhatt2023smoothed} & known & causal & $\log T$ \\
\citet{qian2026} & any & design & $\log T$ \\
\citet{wu2023} & unknown & causal & $(d{+}1)\log^2T$ \\
This paper & unknown & causal & $(d{+}1)\log T$ \\
\bottomrule
\end{tabular}
\caption{Feature access and regret rates in closely related results. ``Causal'' means that features are revealed one at a time; ``design'' means that the full feature sequence is available in advance. The labels are adversarial except in \citet{bhatt2021}, which assumes conditionally Bernoulli labels. The first, second, fifth, and sixth rows concern two-level probability models, whereas \citet{qian2026} study logistic regression. The rates summarize the dependence relevant to this comparison; the protocols and comparator classes are discussed in Appendix~\ref{app:literature}.}
\label{tab:protocols}
\end{table}

Our proof builds on the random-order viewpoint of one-inclusion prediction and the classical Sauer--Shelah trace bounds \citep{haussler1994,sauer1972,shelah1972}. Boolean cubes and related cubical structure have also been studied extensively in the geometry of concept classes and sample compression \citep{rubinstein2009,chalopin2019}. We use the local number of cubes incident to each realizable trace as a probability weight and prove a new coordinate-deletion inequality that controls the resulting unequal splits. For planar halfspaces, this construction can be implemented efficiently using classical halfspace enumeration. This setting concerns two-level linear partitions rather than sigmoid probabilities, and is therefore distinct from efficient improper logistic prediction under norm-dependent guarantees \citep{foster2018,jezequel2020} or with advance access to the feature design \citep{qian2026}. Appendix~\ref{app:literature} gives a more detailed comparison.

\section{Setup and Main Result}\label{sec:setting}

We consider sequential binary prediction over a horizon $T$. At round $t$, the learner observes a feature $X_t\in\Xc$, outputs a probability $\widehat p_t\in(0,1)$ for the event $Y_t=1$, and then observes the binary outcome $Y_t\in\{0,1\}$. The feature sequence is stochastic, whereas the labels may be arbitrary.

A sequential strategy $A$ maps $(x^t,y^{t-1})$ to $\widehat p_t$. Under logarithmic loss,
\[
\ell(p,y)=-y\log p-(1-y)\log(1-p),
\]
its regret relative to a comparator class $\Fc\subseteq[0,1]^{\Xc}$ is
\begin{equation}
R_T(A;x^T,y^T;\Fc)
=\sum_{t=1}^T\ell(\widehat p_t,y_t)
-\inf_{f\in\Fc}\sum_{t=1}^T\ell(f(x_t),y_t).
\label{eq:regretfixed}
\end{equation}
The quantity of interest is the expected worst-case regret~\citep{wu2023}:
\begin{equation}
\widetilde r_T(\Fc)=\inf_A\sup_\mu
\E_{X^T\sim\mu^T}\Bigl[
\sup_{y^T\in\{0,1\}^T}
R_T(A;X^T,y^T;\Fc)
\Bigr].
\label{eq:minimax}
\end{equation}
The order of the quantifiers is important. The strategy is chosen without knowing the feature law $\mu$, while the label sequence may depend on the entire realized feature sequence. Thus the learner must be distribution-blind and the guarantee is uniform over labels.

Let $\Hc\subseteq\{0,1\}^{\Xc}$ be a binary hypothesis class with finite VC dimension $d=\vc(\Hc)$. For a feature prefix $x^t=(x_1,\ldots,x_t)$, define the trace class
\begin{equation}\label{eq:traceclass}
C_t(x^t)
=\{(h(x_1),\ldots,h(x_t)):h\in\Hc\}
\subseteq\{0,1\}^t
\end{equation}
and the growth function
\[
\Pi_{\Hc}(T)=\sup_{x^T\in\Xc^T}|C_T(x^T)|.
\]
We write $\harm_T=\sum_{t=1}^T1/t$ and use natural logarithms throughout. We assume that
\begin{equation}\label{eq:measurable}
\{x^t:v\in C_t(x^t)\}
\end{equation}
is measurable for every $t$ and $v$; Appendix~\ref{app:measurability} verifies this for the standard examples used in the paper.

We will use the following sequential covering notion from~\citep{wu2023}.

\begin{definition}[Stochastic global sequential cover]\label{def:cover}
Fix a class $\Pc$ of laws on $\Xc^T$. A finite family
\[
\Gc\subseteq\{g:\cup_{t=1}^T\Xc^t\to\{0,1\}\}
\]
is an (exact) stochastic global sequential cover of $\Hc$ at failure probability $\delta$ if, for every $P\in\Pc$,
\begin{equation}\label{eq:cover}
P\left\{
\forall h\in\Hc\ \exists g\in\Gc\
\ \forall t\leq T:\;
g(X^t)=h(X_t)
\right\}\geq1-\delta.
\end{equation}
The family $\Gc$ may depend on $\Hc,T,$ and $\delta$, but not on  the feature law $P$.
\end{definition}

The event in \eqref{eq:cover} requires simultaneous coverage of all $h\in\Hc$. Thus the comparator in regret minimization (\ref{eq:minimax}) may be selected after the feature sequence is observed. Moreover, each $g\in\Gc$ is causal: at time $t$ it depends only on $X^t$.

\begin{theorem}[Polynomial stochastic sequential covers]\label{thm:cover}
For every $T\geq1$ and $\delta\in(0,1)$, there is an exact stochastic global sequential cover of $\Hc$ for the class of all exchangeable laws on $\Xc^T$ such that
\begin{equation}\label{eq:coverbound}
\log|\Gc|\leq
6d\harm_T+\log\Pi_{\Hc}(T)+\log(1/\delta)+\log3.
\end{equation}
Consequently,
\begin{equation}\label{eq:coverboundsimple}
\log|\Gc|\leq
7d(1+\log T)+\log(1/\delta)+\log3.
\end{equation}
For fixed $d\geq1$, $T\geq2$, and $\delta=1/T$, one has $|\Gc|=T^{O(d)}$. If $d=0$, a singleton cover suffices.
\end{theorem}

The theorem is stated for exchangeable feature laws, which is slightly stronger than the i.i.d. setting of interest. Exchangeability is sufficient because the proof averages over a random permutation of the observed coordinates, and such a permutation leaves the law unchanged. Compared with \citet[Theorem~4]{wu2023}, our bound removes one factor of $\log T$ and improves the confidence dependence from $\log T\log(1/\delta)$ to $\log(1/\delta)$, thereby resolving the open questions posed in \citet[Remark~4]{wu2023}.

At failure probability $1/T$, Theorem~\ref{thm:cover} gives $\log|\Gc|=O(d\log T)$. Standard aggregation over the resulting finite family then gives logarithmic regret for binary prediction~\citep{wu2023}. In Section~\ref{sec:logalgorithm}, we use the underlying trace distribution directly to obtain the sharper bound for the two-level probability class.

The logarithmic dependence on $T$ in (\ref{eq:coverboundsimple}) cannot be improved in general.

\begin{proposition}[Worst-case lower bound]\label{prop:coverlower}
For every pair of integers $1\leq d\leq T$, there is a binary class $\Hc_d$ of VC dimension $d$ and a nonatomic i.i.d. feature law such that every exact stochastic sequential cover with failure probability $\delta<1$ satisfies
\begin{equation}\label{eq:coverlower}
|\Gc|\geq\sum_{k=0}^{d}\binom{T}{k}.
\end{equation}
Consequently, if $T\geq2d$, then
\[
\log|\Gc|\geq d\log(T/d).
\]
\end{proposition}

For the lower bound, take the class of indicators of subsets of $[0,1]$ of cardinality at most $d$. Under any nonatomic feature law, the $T$ sampled points are distinct almost surely, and the induced trace class contains exactly the binary vectors with at most $d$ ones. A deterministic sequential rule produces only one trace on a fixed sample, so an exact simultaneous cover must contain at least as many rules as there are realizable traces. The proof is given in Appendix~\ref{app:lower}. For fixed $d$, this matches the $O(d\log T)$ upper bound in its dependence on $T$.

\section{Proof of The Covering Theorem}\label{sec:coverproof-main}

We now prove Theorem~\ref{thm:cover}. We assign each realizable trace a local cube weight, use these weights to define sequential transition probabilities, and control the resulting path loss by exposing the coordinates in random reverse order. Arithmetic coding then converts the trace probabilities into a finite deterministic cover. The main argument is given here; technical details are deferred to the appendix.

\subsection{Local cube potential}\label{sec:cubes}

We begin with the local weight that makes the construction sensitive to branch asymmetry. Let $I$ be a finite coordinate set, let $C\subseteq\{0,1\}^I$ be nonempty, and for $v\in C$ and $A\subseteq I$ write $v\oplus\one_A$ for the vector obtained by flipping the coordinates in $A$.

\begin{definition}[Centered-cube weight]\label{def:cubeweight}
For $v\in C$, define
\begin{align}
\Qc_C(v)
&=\{S\subseteq I:v\oplus\one_A\in C\text{ for every }A\subseteq S\},\label{eq:cubes}\\
W_C(v)&=|\Qc_C(v)|.\label{eq:weight}
\end{align}
\end{definition}

A set \(S\in\Qc_C(v)\) specifies a Boolean cube contained in \(C\) with every coordinate outside \(S\) fixed according to \(v\); such cubical structure of concept classes is classical \citep{rubinstein2009,chalopin2019}. In particular, \(S\) is shattered by \(C\), so \(|S|\leq d\) whenever \(\vc(C)\leq d\). Informally, \(W_C(v)\) measures the local combinatorial mass of the class around \(v\): a trace incident to many subcubes retains more local freedom and is therefore safer to bet on. The empty cube always belongs to \(\Qc_C(v)\), so \(W_C(v)\geq1\).

For $j\in I$, let the map $\pi_j$ denote deletion of coordinate $j$, define $C^{-j}=\pi_j(C)$, and write $p=\pi_j(v)$.
 Define
\begin{align}
F_j(p)&=\{u\in C:\pi_j(u)=p\},\label{eq:fiber}\\
R_j(C,v)&=
\frac{\sum_{u\in F_j(p)}W_C(u)}{W_{C^{-j}}(p)},\label{eq:ratio}
\end{align}
Let $v^{(j)}=v\oplus\one_{{j}}$ denote the vector obtained by flipping coordinate $j$. Then the fiber $F_j(p)$ is either $\{v\}$ or $\{v,v^{(j)}\}$, according to whether $v^{(j)}\in C$.

The following lemma is the main combinatorial ingredient of our argument.

\begin{lemma}[Coordinate-deletion inequality]\label{lem:deletion}
If $\vc(C)\leq d$, then, for every $v\in C$,
\begin{equation}\label{eq:deletion}
0<R_j(C,v)\leq4\quad(j\in I),
\qquad
\sum_{j\in I}\pos{R_j(C,v)-1}\leq6d.
\end{equation}
Consequently,
\begin{equation}\label{eq:logdeletion}
\sum_{j\in I}(\log R_j(C,v))_+\leq6d.
\end{equation}
\end{lemma}

The important feature of \eqref{eq:deletion} is that the budget $6d$ is shared across all candidate deletions. This yields a cost of order $d/t$ when $t$ coordinates remain. We give the main calculation here and defer the full proof to Appendix~\ref{app:deletion}. If the fiber $F_j(p)$ contains only $v$, then no centered cube at $v$ contains $j$ and $R_j(C,v)\leq1$. Otherwise the fiber is $\{v,v^{(j)}\}$; writing $B_j(C,v)$ for the number of centered cubes at $v$ that use $j$ and splitting cubes according to whether they use $j$ gives
\begin{equation}\label{eq:keylocal}
W_C(v)+W_C(v^{(j)})
\leq W_{C^{-j}}(p)+3B_j(C,v),
\end{equation}
while projection of cube gives $W_C(v)\leq2W_{C^{-j}}(p)$, thus $(R_j(C,v)-1)_+\leq6B_j(C,v)/W_C(v)$. Summing the incidence count
\begin{equation}\label{eq:incidence}
\sum_{j\in I}B_j(C,v)
=\sum_{S\in\Qc_C(v)}|S|
\leq dW_C(v)
\end{equation}
then yields \eqref{eq:deletion}.

The VC dimension enters only through \eqref{eq:incidence}: each centered cube $S$ contributes once for every coordinate $j\in S$, and $|S|\leq d$ because $S$ is shattered by $C$.

\subsection{Sequential trace probabilities}\label{sec:traceprob}

The deletion inequality is a static statement about a finite trace class. We now turn it into a sequential probability assignment for \emph{realizable} label sequences. The key point is that a branch receives probability in proportion to the local cube mass that survives after taking that branch. Suppose $y^{t-1}\in C_{t-1}(x^{t-1})$, with feasible next labels
\[
A_t(x^t,y^{t-1})
=\{b\in\{0,1\}:y^{t-1}b\in C_t(x^t)\}.
\]
For $b\in A_t(x^t,y^{t-1})$, set
\begin{equation}\label{eq:q}
q_t(b\mid x^t,y^{t-1})
=
\frac{W_{C_t(x^t)}(y^{t-1}b)}
{\sum_{c\in A_t(x^t,y^{t-1})}W_{C_t(x^t)}(y^{t-1}c)}.
\end{equation}
For $b\notin A_t(x^t,y^{t-1})$, set
$q_t(b\mid x^t,y^{t-1})=0$.
The probability assignment is causal and does not involve the feature law.

Note that the probability assignment in \eqref{eq:q} is used only as a splitting weight in the cover construction of Section~\ref{sec:coverproof}; it is \emph{not} itself the predictor used for the hybrid learning problem.

For $h\in\Hc$ write $v_t^h=(h(x_1),\ldots,h(x_t))$ and define the cumulative trace log-loss
\begin{equation}\label{eq:traceloss}
L_T(h;x^T)
=-\sum_{t=1}^T
\log q_t(h(x_t)\mid x^t,v_{t-1}^h).
\end{equation}
The following lemma bounds the cumulative trace log-loss using the cube-weight ratios \eqref{eq:ratio}.
\begin{lemma}\label{lem:telescoping}
Let $w_t=W_{C_t(x^t)}(v_t^h)$, with $w_0=1$, and let $R_t$ be the ratio \eqref{eq:ratio} obtained by deleting the last coordinate of $C_t(x^t)$ at $v_t^h$. Then
\begin{equation}\label{eq:telescoping}
L_T(h;x^T)
=\sum_{t=1}^T\log R_t-\log w_T
\leq\sum_{t=1}^T(\log R_t)_+,
\end{equation}
\end{lemma}

\begin{proof}
Along the trace $v_t^h$, the denominator in \eqref{eq:q} equals $R_t w_{t-1}$, so $q_t=w_t/(R_t w_{t-1})$. Thus
\[
-\log q_t=\log R_t+\log w_{t-1}-\log w_t.
\]
Summing over $t$ telescopes, and the desired inequality follows from $w_T\geq1$.
\end{proof}

\begin{figure}[htbp]
\centering
\begin{tikzpicture}[
  x=1cm,y=0.82cm,font=\small,
  trace/.style={draw=black!60,rounded corners=2pt,fill=white,
    minimum width=1.2cm,minimum height=0.75cm,align=center,inner sep=3pt},
  focus/.style={trace,draw=figblue,fill=figlight,line width=0.8pt},
  edge/.style={draw=black!65,line width=0.7pt},
  split/.style={-{Stealth[length=2.2mm]},draw=figblue,line width=1pt}
]
\path[use as bounding box] (0,-0.65) rectangle (15.8,4.15);
\node[font=\small\bfseries] at (3.5,3.95) {(a) Local cube weights};
\node[font=\small\bfseries] at (11.75,3.95) {(b) Weighted split at round 3};

\node[focus] (zero) at (3.5,1.2) {$000$\\[-1pt]$W=4$};
\node[trace] (first) at (3.5,3.0) {$100$\\[-1pt]$W=2$};
\node[trace] (second) at (1.25,0.2) {$010$\\[-1pt]$W=2$};
\node[trace] (third) at (5.75,0.2) {$001$\\[-1pt]$W=2$};
\draw[edge] (zero) -- (first);
\draw[edge] (zero) -- (second);
\draw[edge] (zero) -- (third);
\node[font=\footnotesize] at (3.5,-0.85)
  {$W(000)=1\text{ vertex}+3\text{ edges}=4$};

\node[trace] (prefix) at (11.75,2.85) {{\footnotesize prefix}\\[-1pt]$00$};
\node[focus] (childzero) at (9.6,0.45) {$000$\\[-1pt]$W=4$};
\node[trace] (childone) at (13.9,0.45) {$001$\\[-1pt]$W=2$};
\draw[split] (prefix.south) -- (childzero.north);
\draw[split] (prefix.south) -- (childone.north);
\node[anchor=east,text=figblue] at (10.45,1.9)
  {$\displaystyle\frac{4}{4+2}=\frac23$};
\node[anchor=west,text=figblue] at (13.05,1.9)
  {$\displaystyle\frac{2}{4+2}=\frac13$};
\node[font=\footnotesize] at (11.75,-0.6)
  {Children receive mass in proportion to their weights.};
\end{tikzpicture}
\caption{Cube weights and asymmetric branching for
$C_3=\{000,100,010,001\}$. (a) Each weight counts the vertex and its incident edges;
no higher-dimensional cubes occur. (b) The extensions of prefix $00$
receive probabilities $2/3$ and $1/3$ under \eqref{eq:q}.}
\label{fig:cube-split}
\end{figure}
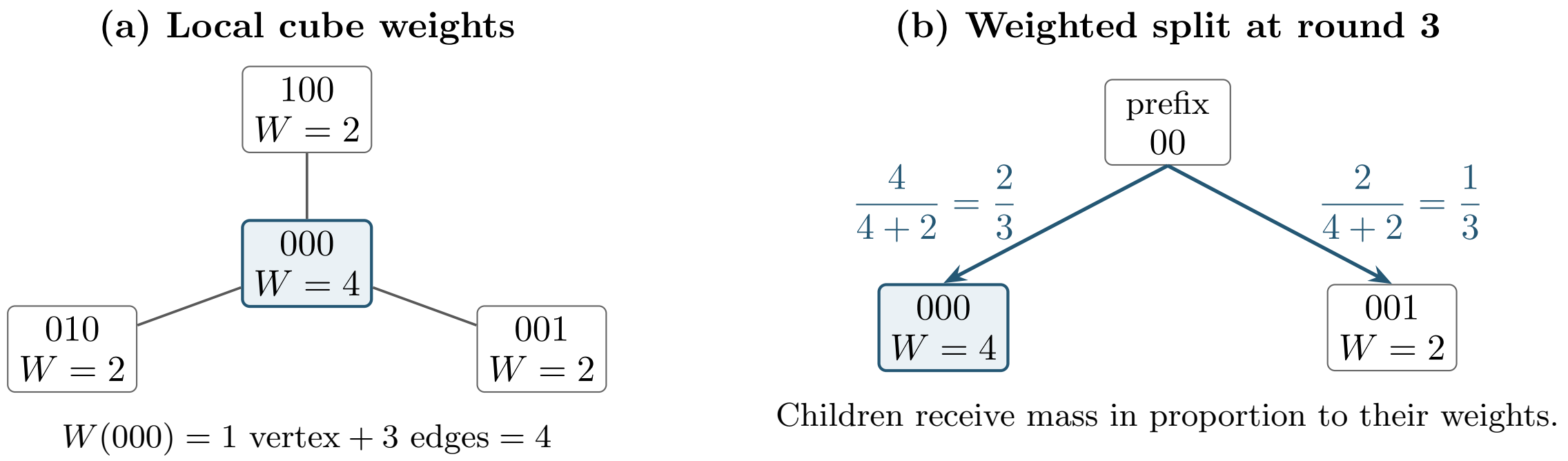

\begin{example}
Take the class consisting of the zero function and all singleton indicators, whose traces on distinct features are $C_t=\{0^t,e_1,\ldots,e_t\}$. Its VC dimension is one. The zero trace has $t$ single-coordinate flips that remain in $C_t$, so $W_{C_t}(0^t)=t+1$, whereas every other trace has weight $2$. Along the all-zero path, both extensions remain feasible at every round,
but the cube-weighted split is highly asymmetric:
$$q_t(0\mid x^t,0^{t-1})=\frac{t+1}{t+3}.$$ Hence the probability assigned to
the all-zero trace over $T$ rounds is
$6/((T+2)(T+3))$, and its cumulative trace log-loss is only $O(\log T)$. In
contrast, splitting equally whenever both extensions are feasible would
assign probability $2^{-T}$ and incur log-loss $T\log 2$.
Figure~\ref{fig:cube-split} illustrates the weights and the split at $t=3$.
\end{example}

\FloatBarrier

\subsection{Random-order control of realizable traces}\label{sec:randomorder-main}

The bound in Lemma~\ref{lem:telescoping} holds pointwise for every realizable trace. We now convert it into a uniform high-probability bound under exchangeable feature sequences. This uniform trace guarantee is the probabilistic heart of our arguments.

\begin{proposition}[Uniform trace log-loss]\label{prop:traceloss}
Let $P$ be any exchangeable law on $\Xc^T$. Then
\begin{equation}\label{eq:expectedtrace}
\E_P\left[\sup_{h\in\Hc}L_T(h;X^T)\right]
\leq6d\harm_T+\log\Pi_{\Hc}(T),
\end{equation}
and for every $\delta\in(0,1)$,
\begin{equation}\label{eq:tailtrace}
P\Bigl\{\sup_{h\in\Hc}L_T(h;X^T)
>6d\harm_T+\log\Pi_{\Hc}(T)+\log(1/\delta)\Bigr\}
\leq\delta.
\end{equation}
\end{proposition}

\begin{proof}
Fix a deterministic sample $x^T$ and a full trace $v\in C_T(x^T)$. Let $\sigma$ be a uniformly random permutation of $[T]$ and put $S_t=\{\sigma(1),\ldots,\sigma(t)\}$. Write $C_{S_t}$ for the projection of $C_T(x^T)$ onto $S_t$ and $v_{S_t}$ for the corresponding projection of $v$, and define
\[
Y_t=\bigl(\log R_{\sigma(t)}(C_{S_t},v_{S_t})\bigr)_+.
\]
Let $\mathcal F_t$ be the sigma-field generated by the suffix $(\sigma(t+1),\ldots,\sigma(T))$, with $\mathcal F_T$ trivial. Here, we may view the permutation as being revealed \emph{backward}, from $\sigma(T)$ to $\sigma(1)$.
Conditional on $\mathcal F_t$, the set $S_t$ is fixed and $\sigma(t)$ is uniform on $S_t$. Therefore Lemma~\ref{lem:deletion} gives
\begin{align}
\E[e^{Y_t}\mid\mathcal F_t]
&=\frac1t\sum_{j\in S_t}\exp\!\left((\log R_j(C_{S_t},v_{S_t}))_+\right)\notag\\
&=1+\frac1t\sum_{j\in S_t}(R_j(C_{S_t},v_{S_t})-1)_+
\leq 1+\frac{6d}{t}.\label{eq:onestepmgf}
\end{align}
Iterating conditional expectations,
\begin{equation}\label{eq:mgfall}
\E_\sigma\exp\left(\sum_{t=1}^TY_t\right)
\leq\prod_{t=1}^T\left(1+\frac{6d}{t}\right)
\leq e^{6d\harm_T}.
\end{equation}

By Lemma~\ref{lem:telescoping}, the cumulative trace log-loss in the permuted order is at most $\sum_tY_t$. Hence, for every fixed terminal trace $v$,
\[
\E_\sigma e^{L_T(v;x^T_\sigma)}\leq e^{6d\harm_T}.
\]
There are at most $|C_T(x^T)|\leq\Pi_{\Hc}(T)$ realizable terminal traces. Since
\[
e^{\max_v L_T(v;x^T_\sigma)}
\leq\sum_{v\in C_T(x^T)}e^{L_T(v;x^T_\sigma)},
\]
we obtain
\begin{equation}\label{eq:maxmgf-main}
\E_\sigma e^{\max_v L_T(v;x^T_\sigma)}
\leq \Pi_{\Hc}(T)e^{6d\harm_T}.
\end{equation}
Taking logarithms and Jensen's inequality gives
\[
\E_\sigma\max_vL_T(v;x^T_\sigma)
\leq6d\harm_T+\log\Pi_{\Hc}(T),
\]
while Markov's inequality applied to \eqref{eq:maxmgf-main} gives, for every $\delta\in(0,1)$,
\[
\Pr_\sigma\Bigl\{
\max_vL_T(v;x^T_\sigma)
>6d\harm_T+\log\Pi_{\Hc}(T)+\log(1/\delta)
\Bigr\}
\leq\delta.
\]

Finally let $X^T$ have an exchangeable law and let $\sigma$ be independent and uniform. Then $X^T_\sigma\stackrel{d}=X^T$. Averaging the two preceding bounds over $X^T$ therefore proves \eqref{eq:expectedtrace} and \eqref{eq:tailtrace}, hence Proposition~\ref{prop:traceloss}.
\end{proof}

The reverse exposure of the permutation is used only for the analysis; the learner still observes the features in their original order. Its role is to make the next deleted coordinate uniform among those that remain. The exponential-moment bound then allows us to control all realizable traces simultaneously at an additional cost of only $\log\Pi_{\Hc}(T)$. Appendix~\ref{app:randomorder} provides the full permutation bookkeeping.
\subsection{Constructing the sequential cover via arithmetic coding}\label{sec:coverproof}

Proposition~\ref{prop:traceloss} gives a uniform lower bound on the probability assigned to every realizable trace. It remains to convert this probabilistic guarantee into a fixed finite family of deterministic sequential rules. For any fixed feature sequence, the probabilities in \eqref{eq:q} define a binary probability tree over the realizable traces. We construct a finite family of sequential rules whose paths cover this tree using an interval construction reminiscent of arithmetic coding \citep{witten1987}.

Assign the root of the probability tree the interval
$I_{\varnothing}=[0,1)$. Recursively, suppose a realizable history
$y^{t-1}$ has been assigned an interval $I_{y^{t-1}}$. Partition this
interval into two consecutive subintervals $I_{y^{t-1}0}$ and
$I_{y^{t-1}1}$ whose lengths are, respectively,
$q_t(0\mid x^t,y^{t-1})$ and $q_t(1\mid x^t,y^{t-1})$ times the
length of the parent interval.
For any realizable trace
$v_T^h=(h(x_1),\ldots,h(x_T))$, the length of its terminal interval is
the product of the transition probabilities along the path:
\begin{equation}\label{eq:intervallength}
|I_{v_T^h}|
=
\prod_{t=1}^T
q_t(h(x_t)\mid x^t,v_{t-1}^h)
=
e^{-L_T(h;x^T)}.
\end{equation}

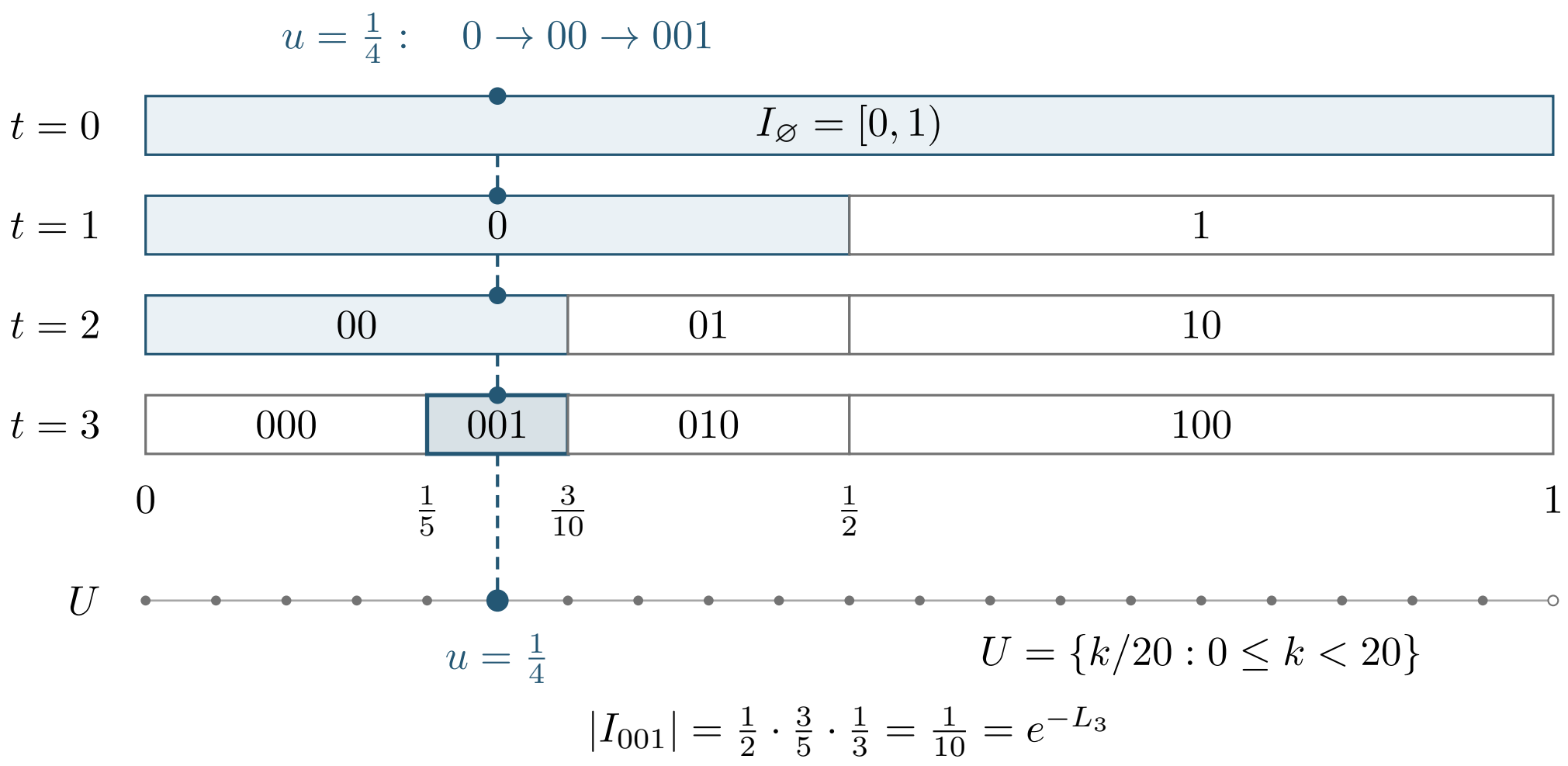
\begin{figure}[htbp]
\centering
\begin{tikzpicture}[
  x=1cm,y=1cm,font=\small,
  interval/.style={draw=black!55,line width=.6pt,fill=white},
  selected/.style={interval,fill=figlight,draw=figblue},
  pathmark/.style={circle,fill=figblue,inner sep=1.5pt},
  seedpath/.style={figblue,densely dashed,line width=.8pt}
]
  \node[anchor=east] at (-.25,3.4) {$t=0$};
  \node[anchor=east] at (-.25,2.55) {$t=1$};
  \node[anchor=east] at (-.25,1.7) {$t=2$};
  \node[anchor=east] at (-.25,.85) {$t=3$};

  \path[selected] (0,3.15) rectangle (12,3.65);
  \node at (6,3.4) {$I_{\varnothing}=[0,1)$};

  \path[selected] (0,2.3) rectangle (6,2.8);
  \path[interval] (6,2.3) rectangle (12,2.8);
  \node at (3,2.55) {$0$};
  \node at (9,2.55) {$1$};

  \path[selected] (0,1.45) rectangle (3.6,1.95);
  \path[interval] (3.6,1.45) rectangle (6,1.95);
  \path[interval] (6,1.45) rectangle (12,1.95);
  \node at (1.8,1.7) {$00$};
  \node at (4.8,1.7) {$01$};
  \node at (9,1.7) {$10$};

  \path[interval] (0,.6) rectangle (2.4,1.1);
  \path[selected,fill=figblue!18,line width=1pt] (2.4,.6) rectangle (3.6,1.1);
  \path[interval] (3.6,.6) rectangle (6,1.1);
  \path[interval] (6,.6) rectangle (12,1.1);
  \node at (1.2,.85) {$000$};
  \node at (3,.85) {$001$};
  \node at (4.8,.85) {$010$};
  \node at (9,.85) {$100$};

  \foreach \top/\bottom in {3.15/2.8,2.3/1.95,1.45/1.1}
    \draw[seedpath] (3,\top) -- (3,\bottom);
  \foreach \yy in {3.65,2.8,1.95,1.1}
    \node[pathmark] at (3,\yy) {};
  \node[anchor=south,text=figblue] at (3,3.8) {$u=\tfrac14:\quad 0\to00\to001$};

  \foreach \xx/\label in {0/0,2.4/\frac15,3.6/\frac3{10},6/\frac12,12/1}
    \node[anchor=north] at (\xx,.46) {$\label$};

  \node[anchor=east,align=right] at (-.25,-.65) {$U$};
  \draw[black!35,line width=.5pt] (0,-.65) -- (12,-.65);
  \foreach \k in {0,...,19}
    \fill[black!55] ({.6*\k},-.65) circle (1.2pt);
  \draw[black!55,fill=white] (12,-.65) circle (1.2pt);
  \draw[seedpath] (3,.6) -- (3,-.65);
  \fill[figblue] (3,-.65) circle (2.7pt);
  \node[anchor=north,text=figblue] at (3,-.8) {$u=\tfrac14$};
  \node[anchor=north] at (9,-.8) {$U=\{k/20:0\leq k<20\}$};
  \node[anchor=north] at (6,-1.42)
    {$|I_{001}|=\tfrac12\cdot\tfrac35\cdot\tfrac13=\tfrac1{10}=e^{-L_3}$};
\end{tikzpicture}
\caption{Arithmetic intervals for $C_3=\{000,100,010,001\}$. Each row partitions $[0,1)$ into the intervals of feasible prefixes. Shading marks the nested intervals containing the fixed seed $u=1/4$, whose rule predicts $0,0,1$. The terminal interval has length equal to the product of its path probabilities. The illustrative grid has $M=20$; all terminal intervals have length at least $1/10=2/M$ and therefore contain grid points. In the general construction, $M=\lceil2e^{L_*}\rceil$. All intervals are left-closed and right-open.}
\label{fig:arithmetic-cover}
\end{figure}

Writing
\[
L_*=6d\harm_T+\log\Pi_{\Hc}(T)+\log(1/\delta),
\]
take the fixed grid
\[
U=\left\{\frac{k}{M}:k=0,\ldots,M-1\right\},
\qquad M=\lceil2e^{L_*}\rceil.
\]
Each grid point $u\in U$ defines a deterministic sequential rule $g_u$: starting from the root, at each round $g_u$ follows the unique child interval containing $u$ and predicts the label associated with that child.
Note that $g_u$ is sequential: at round $t$, its prediction depends only on the observed feature prefix $X^t$.
Figure~\ref{fig:arithmetic-cover} illustrates the construction for the same
singleton class as Figure~\ref{fig:cube-split}.

\begin{proof}[Proof of Theorem~\ref{thm:cover}]
Let $E$ denote the event in Proposition~\ref{prop:traceloss} on which
\[
\sup_{h\in\Hc}L_T(h;X^T)\leq L_*.
\]
By \eqref{eq:tailtrace}, every exchangeable feature law satisfies $P(E)\geq1-\delta$.

Fix a realization $x^T\in E$ and any $h\in\Hc$. The nested interval associated with the trace
\[
(h(x_1),\ldots,h(x_T))
\]
has terminal length, by \eqref{eq:intervallength},
\[
e^{-L_T(h;x^T)}\geq e^{-L_*}.
\]
Since $M\geq2e^{L_*}$, every interval in $[0,1)$ of length at least $e^{-L_*}$ contains a point of $U$; Appendix~\ref{app:arithmetic} gives the elementary boundary check. Choose such a point $u$. Because the intervals are nested, the rule $g_u$ follows the target branch at every prefix, and therefore
\[
g_u(x^t)=h(x_t),\qquad t=1,\ldots,T.
\]
The same grid $U$ is fixed before the feature law is chosen, and the argument holds simultaneously for every $h$. Thus
\[
\Gc=\{g_u:u\in U\}
\]
is an exact distribution-blind stochastic global sequential cover on $E$.

It remains only to bound its size. Since $L_*\geq0$,
\[
|\Gc|\leq M
\leq 2e^{L_*}+1
\leq3e^{L_*},
\]
and hence
\[
\log|\Gc|
\leq6d\harm_T+\log\Pi_{\Hc}(T)
+\log(1/\delta)+\log3,
\]
which is \eqref{eq:coverbound}. Sauer--Shelah gives
\[
\Pi_{\Hc}(T)\leq\sum_{k=0}^{d}\binom Tk
\leq (eT/d)^d
\]
when $1\leq d\leq T$. If $d>T$, then
$\log\Pi_{\Hc}(T)\leq T\log2\leq d(1+\log T)$.
Together with $\harm_T\leq1+\log T$, these bounds give
\eqref{eq:coverboundsimple} for every $d\geq1$.
If $d=0$, all hypotheses agree pointwise, and a singleton cover suffices.
This proves Theorem~\ref{thm:cover}.
\end{proof}

\section{Application to Log-loss}
\label{sec:logalgorithm}
\label{sec:planar}

We now give an immediate consequence of Theorem~\ref{thm:cover} for log-loss regrets; see \citet{wu2023} for further applications of stochastic sequential covers.

Consider the two-level probability class
\begin{equation}
\label{eq:twolevel}
\Fc_{\Hc}
=
\left\{
x\mapsto a+(b-a)h(x):
h\in\Hc,\ a,b\in[0,1]
\right\},
\end{equation}
where $\Hc\subseteq\{0,1\}^{\Xc}$ is a binary-valued class. Here, $h$
partitions the feature space into two regions, while $a$ and $b$ specify
the probability of label $1$ on the two regions. This class has been studied in \citet{bhatt2021,bilodeau2023density,wu2023}.

\begin{theorem}[Logarithmic regret]
\label{thm:logloss}
For $T\geq2$,
\begin{equation}
\label{eq:loglossbound}
\widetilde r_T(\Fc_{\Hc})
\leq
\left(1+\frac1T\right)
\Bigl(
6d\harm_T+\log\Pi_{\Hc}(T)+3\log T+\log6
\Bigr)+3.
\end{equation}
In particular,
\[
\widetilde r_T(\Fc_{\Hc})
=
O((d+1)\log T).
\]
\end{theorem}

\begin{proof}
Apply Theorem~\ref{thm:cover} with $\delta=1/T$, and let $\Gc$ be the resulting exact cover of $\Hc$. Let
\[
\Theta_T=\left\{0,\frac1{T-1},\frac{2}{T-1},\ldots,1\right\},
\]
so $|\Theta_T|=T$ and every point of $[0,1]$ is within $1/T$ of some element of $\Theta_T$. For $g\in\Gc$ and $\bar a,\bar b\in\Theta_T$, define
\[
g_{\bar a,\bar b}(x^t)
=
\bar a+(\bar b-\bar a)g(x^t),
\]
and let
\[
\Gc'
=
\{g_{\bar a,\bar b}:g\in\Gc,\ \bar a,\bar b\in\Theta_T\}.
\]
Then $|\Gc'|\leq T^2|\Gc|$.
On the success event of $\Gc$, every $f(x)=a+(b-a)h(x)\in\Fc_{\Hc}$ is uniformly approximated along the realized sequence to accuracy $1/T$ by some member of $\Gc'$. Hence $\Gc'$ is a stochastic sequential $1/T$-cover of $\Fc_{\Hc}$ with failure probability $1/T$. By \citet[Theorem~2]{wu2023},
\[
\widetilde r_T(\Fc_{\Hc})
\leq
3+\left(1+\frac1T\right)\log(|\Gc'|+1).
\]
Since $|\Gc'|\leq T^2|\Gc|$ and $|\Gc|\geq1$,
\[
\log(|\Gc'|+1)
\leq
2\log T+\log|\Gc|+\log2.
\]
Applying Theorem~\ref{thm:cover} with $\delta=1/T$ yields \eqref{eq:loglossbound}.
\end{proof}

Theorem~\ref{thm:logloss} achieves the \emph{optimal} logarithmic dependence on $T$ for general VC classes, essentially closing the line of work initiated in \citet{bhatt2021} and further developed in \citet{bilodeau2023density,wu2023,bhatt2023portfolio}.

A naive implementation of the predictor achieving the regret bound in Theorem~\ref{thm:logloss} aggregates over the cover $\Gc'$, whose size is $T^{O(d+1)}$. For fixed $d$, this gives polynomially many experts; a polynomial-time implementation additionally requires efficient construction and evaluation of those experts. Finite VC dimension alone does not supply this computational access. For planar linear classifiers, using the probability assignment in \eqref{eq:q}, we construct the predictor directly without explicitly aggregating over the cover. The full construction and analysis are given in Appendix~\ref{app:mixture}. For any $\Hc$ and every exchangeable feature law, the direct predictor satisfies the sharper bound
\begin{equation}
\label{eq:direct-mixture-bound}
\E\left[
\sup_{y^T}
R_T(A;X^T,y^T;\Fc_{\Hc})
\right]
\leq
6d\harm_T+\log\Pi_{\Hc}(T)+2\log(T+1).
\end{equation}

\begin{theorem}[Efficient prediction for planar halfspaces]
\label{thm:planar}
Let $\Hc$ be the class of affine halfspaces in $\R^2$. There exists a
distribution-blind sequential predictor for $\Fc_{\Hc}$ with
$O(\log T)$ expected worst-case log-loss regret under unknown i.i.d.\
features. At round $t$, the predictor maintains at most
$t^2-t+2$ realizable labelings of the observed feature prefix. For
rational features, it admits an exact implementation with polynomial
bit complexity, using $O(T^6)$ combinatorial and rational-arithmetic
operations over $T$ rounds.
\end{theorem}

Indeed, affine halfspaces in $\R^2$ realize at most
$t^2-t+2=O(t^2)$ labelings on $t$ points. Since their VC dimension is
three, each cube weight can be computed by examining $O(t^3)$ candidate
coordinate sets. Thus one round requires $O(t^5)$ operations, giving
$O(T^6)$ over the full horizon. Appendix~\ref{app:planar} gives the
implementation and bit-complexity details.

\section{Limitation of Pseudo-Dimension}\label{sec:pseudo}

The binary result naturally raises the question of whether the same
phenomenon extends to real-valued classes, with VC dimension replaced by
pseudo-dimension. For $\Fc\subseteq[0,1]^{\Xc}$, the pseudo-dimension is
the VC dimension of the subgraph class
\[
\left\{
(x,r)\mapsto \one\{f(x)\geq r\}: f\in\Fc
\right\}.
\]
The corresponding sequential covering notion extends to $[0,1]$-valued
predictors and requires
\[
|g(X^t)-f(X_t)|\leq\alpha
\]
on every round~\citep{wu2023}.

It was shown in \citet[Appendix~E]{wu2023} that a real-valued class $\Fc$ of pseudo-dimension $d$ admits a stochastic sequential $\alpha$-cover at failure probability $\delta$ with log-cardinality
\[
O\!\left(d\log^2(T/\alpha)+\log(T/\alpha)\log(1/\delta)\right).
\]
For log loss, taking $\alpha=\delta=1/T$ yields an $O(d\log^2 T)$ regret bound for every such class with $d\geq1$. Quite surprisingly, our next theorem shows that, unlike in the binary case, this extra logarithmic factor \emph{cannot} in general be removed: even for a pseudo-dimension-one class, a stochastic sequential $1/T$-cover with log-cardinality $O(\log T)$ is \emph{impossible}.

\begin{theorem}[Pseudo-dimension does not imply polynomial covers]\label{thm:pdim}
There is a fixed class $\Fc_*\subseteq(0,1)^{\Xc}$ with $\pdim(\Fc_*)=1$ and universal constants $c,T_0>0$ such that, for every $T\geq T_0$, every distribution-blind stochastic sequential $1/T$-cover of $\Fc_*$ at failure probability $1/4$ satisfies
\begin{equation}\label{eq:pdimbound}
\log|\Gc|\geq c\log^2T.
\end{equation}
The class can be taken to be the sigmoid family
\begin{equation}\label{eq:sigmoidclass}
\Fc_*=\{(a,b)\mapsto \sigma(a\theta+b):\theta\in(0,1)\},
\qquad \sigma(s)=\frac1{1+e^{-s}},
\end{equation}
on the domain $(a,b)\in(0,\infty)\times\R$.
\end{theorem}

We sketch the key idea below and refer to Appendix~\ref{app:pdimlower} for the complete proof.

Let $K\asymp1/\alpha$. We construct a full $K$-ary tree in which each node $u$ is associated with a parameter interval $I_u\subset(0,1)$ and a feature $x_u$. Its children $I_{u1},\ldots,I_{uK}$ are disjoint subintervals of $I_u$ chosen so that, whenever $\theta\in I_{uk}$, the value $f_\theta(x_u)$ lies in one of $K$ output bands separated by more than $2\alpha$. Therefore, any sequential predictor that approximates $f_\theta(x_u)$ within accuracy $\alpha$ uniquely identifies which child interval $I_{uk}$ contains $\theta$. The key idea is thus to reduce the covering problem to identifying a hidden root-to-leaf path in this tree.

We randomize the hard instance in two stages. First choose a hidden path of depth $L$
\[
B=(B_1,\ldots,B_L),
\qquad
B_1,\ldots,B_L
\stackrel{\mathrm{iid}}{\sim}\operatorname{Unif}[K],
\]
and choose $\theta_B$ in the corresponding terminal interval
$I_{B_1\cdots B_L}$. The node at depth $r$ on this path has feature
\[
x_r(B)=x_{B_1\cdots B_{r-1}}.
\]
For this fixed path $B$, let $\mu_B$ be the uniform distribution on
$x_1(B),\ldots,x_L(B)$, and draw features
\[
X_1,\ldots,X_T
\stackrel{\mathrm{iid}}{\sim}\mu_B.
\]
Equivalently, we can draw independent depths
\[
D_1,\ldots,D_T
\stackrel{\mathrm{iid}}{\sim}\operatorname{Unif}[L]
\]
and set $X_t=x_{D_t}(B)$. Thus $B$ determines the hidden path and target
function, while $D_t$ determines which node on that fixed path is sampled
at round $t$. Conditional on $B$, the features are i.i.d.

Call round $t$ a \emph{record} if
\[
D_t>\max_{s<t}D_s,
\]
with the first round counted as a record. Suppose $D_t=r$ is a record.
Since all previously sampled depths are smaller than $r$, the feature
prefix $X^t$ depends only on $B_1,\ldots,B_{r-1}$. Any sequential rule
must therefore choose its prediction before learning the fresh digit
$B_r$. On the other hand, $f_{\theta_B}(X_t)$ lies in the output band
indexed by $B_r$. Because the bands are separated by more than
$2\alpha$, predicting within $\alpha$ uniquely identifies $B_r$.
Hence any fixed sequential rule succeeds at this record with conditional
probability at most $1/K$.

Different records expose fresh path digits. Therefore, if $R$ records
occur, any fixed sequential rule tracks the target with probability at
most $K^{-R}$. For a cover of size $M$, a union bound gives
\begin{equation}
\label{eq:lowerprob}
\frac34
\leq
\E\min\{1,MK^{-R}\}.
\end{equation}

As $L\to\infty$, collisions among the sampled depths vanish, and the
relative order of $D_1,\ldots,D_T$ converges to that of a uniformly
random permutation of $[T]$. Thus $R$ converges to the corresponding
record count, which is of order $\log T$. Combining
\eqref{eq:lowerprob} with the generating function of the record count
yields
\[
\log M
=
\Omega(\log T\log K)
=
\Omega\left(\log T\log\frac1\alpha\right).
\]
Taking $\alpha=1/T$ gives the claimed $\Omega(\log^2 T)$ lower bound.
Appendix~\ref{app:pdimlower} gives the full construction and proof,
including the conditioning argument and the record-count calculation.

\section{Discussion}
We established tight stochastic sequential covering bounds for general
finite-VC classes, a key ingredient for sharp regret guarantees in
distribution-blind hybrid online learning. For the two-level
probability class under log loss, the new covering bound yields the
optimal logarithmic dependence on the horizon, and similar reductions
may improve other prediction problems, including mixable losses and
contextual bandits. Our pseudo-dimension lower bound shows that the
binary theory does not extend automatically to real-valued classes.
For structured models such as logistic regression with unknown
i.i.d.\ features, it remains open whether dimension alone suffices for
$O(d\log T)$ distribution-blind regret.

\clearpage
\section*{AI Use Statement}
Generative AI tools were used in the development of this work. In particular, OpenAI GPT-5.6 Sol provided the key cube-potential argument underlying the main covering theorem, including the idea of weighting traces by their incident Boolean subcubes and using these weights for unequal probability splitting. The authors formulated the research problem, supplied the relevant prior results and partial progress, directed and iteratively steered the exploration, and independently checked the resulting mathematical arguments and proofs. Anthropic Claude and OpenAI Codex were also used to help investigate extensions, search and organize relevant literature, revise exposition, implement verification code and predictors, and check the compiled manuscript. %
\bibliography{references}

\clearpage
\appendix
\section{Proof of the Coordinate-Deletion Inequality}\label{app:deletion}

We prove Lemma~\ref{lem:deletion}. Fix a finite coordinate set $I$, a nonempty class $C\subseteq\{0,1\}^I$, a vertex $v\in C$, and a coordinate $j\in I$. Let
\[
D=W_{C^{-j}}(p),
\qquad p=\pi_j(v),
\]
and let $B=B_j(C,v)$ denote the number of members of $\Qc_C(v)$ containing $j$.

If $F_j(p)=\{v\}$, then $v\oplus\one_{\{j\}}\notin C$. Hence no centered cube at $v$ contains $j$. Deleting coordinate $j$ maps every centered cube at $v$ injectively to a centered cube at $p$ in $C^{-j}$. Consequently,
\[
W_C(v)\leq D,
\qquad
R_j(C,v)\leq1,
\qquad
B=0.
\]

Suppose now that
\[
F_j(p)=\{v,v^{(j)}\},
\qquad
v^{(j)}=v\oplus\one_{\{j\}}.
\]
For $S\subseteq I\setminus\{j\}$, define
\[
a_S=\one\{S\in\Qc_C(v)\},
\qquad
b_S=\one\{S\in\Qc_C(v^{(j)})\},
\qquad
c_S=\one\{S\in\Qc_{C^{-j}}(p)\}.
\]
A cube on $S\cup\{j\}$ is present at $v$ if and only if the complete $S$-cube is present in both $j$-layers. The same equivalence holds at $v^{(j)}$. Therefore,
\begin{align}
W_C(v)&=\sum_{S\subseteq I\setminus\{j\}}(a_S+a_Sb_S),\label{eq:appendix-wv}\\
W_C(v^{(j)})&=\sum_{S\subseteq I\setminus\{j\}}(b_S+a_Sb_S),\label{eq:appendix-wvj}\\
B&=\sum_{S\subseteq I\setminus\{j\}}a_Sb_S.\label{eq:appendix-B}
\end{align}
Whenever $a_S=1$ or $b_S=1$, the projected $S$-cube is present in $C^{-j}$, and hence $a_S,b_S\leq c_S$. For the four possible pairs $(a_S,b_S)$,
\[
a_S+b_S+2a_Sb_S\leq c_S+3a_Sb_S.
\]
Summing and using \eqref{eq:appendix-wv}--\eqref{eq:appendix-B} gives
\begin{equation}\label{eq:appendix-numerator}
W_C(v)+W_C(v^{(j)})\leq D+3B.
\end{equation}
Every cube counted by $B$ projects to a centered cube at $p$, so $B\leq D$. It follows from \eqref{eq:appendix-numerator} that
\[
R_j(C,v)\leq1+3B/D\leq4.
\]
Moreover, \eqref{eq:appendix-wv} and $a_S,a_Sb_S\leq c_S$ imply
\[
W_C(v)\leq2D.
\]
Thus, in both fiber cases,
\begin{equation}\label{eq:appendix-pointwise}
(R_j(C,v)-1)_+
\leq\frac{3B_j(C,v)}{W_{C^{-j}}(p)}
\leq\frac{6B_j(C,v)}{W_C(v)}.
\end{equation}
Finally, double counting cube--coordinate incidences gives
\[
\sum_{j\in I}B_j(C,v)
=\sum_{j\in I}\sum_{S\in\Qc_C(v)}\one\{j\in S\}
=\sum_{S\in\Qc_C(v)}|S|
\leq d|\Qc_C(v)|
=dW_C(v),
\]
because every $S\in\Qc_C(v)$ is shattered and therefore has cardinality at most $d$. Summing \eqref{eq:appendix-pointwise} proves
\[
\sum_{j\in I}(R_j(C,v)-1)_+\leq6d.
\]
The logarithmic statement follows from
\[
(\log r)_+\leq(r-1)_+,
\qquad r>0.
\]

\section{Random-Order Analysis}\label{app:randomorder}

We prove Proposition~\ref{prop:traceloss}. Fix a deterministic feature sequence $x^T=(x_1,\ldots,x_T)$ and a full trace
\[
v=(h(x_1),\ldots,h(x_T))\in C_T(x^T).
\]
For $S\subseteq[T]$, define
\[
C_S=\{(h(x_i))_{i\in S}:h\in\Hc\},
\qquad
v_S=v|_S.
\]
If $j\in S$, let
\[
R(S,j;v_S)
=
\frac{\sum_{u\in C_S:u|_{S\setminus\{j\}}=v_{S\setminus\{j\}}}W_{C_S}(u)}
{W_{C_{S\setminus\{j\}}}(v_{S\setminus\{j\}})}.
\]
This is precisely the ratio in \eqref{eq:ratio} for the class $C_S$ and coordinate $j$.

Let $\sigma$ be a uniformly random permutation of $[T]$, and set
\[
S_t=\{\sigma(1),\ldots,\sigma(t)\}.
\]
When the features are presented in the order $x_{\sigma(1)},\ldots,x_{\sigma(T)}$, the forward ratio at time $t$ is
\[
R_t^\sigma=R(S_t,\sigma(t);v_{S_t}).
\]
Define
\[
Y_t=(\log R_t^\sigma)_+.
\]

Expose the permutation in the reverse order
\[
\sigma(T),\sigma(T-1),\ldots,\sigma(1).
\]
Let $\mathcal F_t$ be the sigma-algebra generated by the already exposed suffix $\sigma(T),\ldots,\sigma(t+1)$. Before $\sigma(t)$ is exposed, the set $S_t$ is known and $\sigma(t)$ is uniform on $S_t$, even conditional on $\mathcal F_t$. Lemma~\ref{lem:deletion} therefore gives
\begin{align}
\E_\sigma[e^{Y_t}\mid\mathcal F_t]
&=\frac1t\sum_{j\in S_t}\exp\bigl((\log R(S_t,j;v_{S_t}))_+\bigr)\\
&=\frac1t\sum_{j\in S_t}\max\{1,R(S_t,j;v_{S_t})\}\\
&=1+\frac1t\sum_{j\in S_t}(R(S_t,j;v_{S_t})-1)_+\\
&\leq1+\frac{6d}{t}.\label{eq:appendix-mgfstep}
\end{align}
In the reverse exposure order, the factors corresponding to already exposed coordinates are measurable. Iterated conditional expectation and \eqref{eq:appendix-mgfstep} yield
\begin{equation}\label{eq:appendix-mgf}
\E_\sigma\exp\left(\sum_{t=1}^TY_t\right)
\leq\prod_{t=1}^T\left(1+\frac{6d}{t}\right)
\leq e^{6d\harm_T}.
\end{equation}
By Lemma~\ref{lem:telescoping},
\[
L_T(v;\sigma)\leq\sum_{t=1}^TY_t.
\]
Consequently,
\begin{equation}\label{eq:appendix-fixedmgf}
\E_\sigma e^{L_T(v;\sigma)}\leq e^{6d\harm_T}.
\end{equation}
Markov's inequality implies that, for every $\varepsilon\in(0,1)$,
\begin{equation}\label{eq:appendix-fixedtail}
\Prb_\sigma\left\{L_T(v;\sigma)>6d\harm_T+\log(1/\varepsilon)\right\}
\leq\varepsilon.
\end{equation}

Let
\[
N(x^T)=|C_T(x^T)|.
\]
There are only $N(x^T)$ distinct full traces, even if $\Hc$ is infinite. Applying \eqref{eq:appendix-fixedtail} with $\varepsilon=\delta/N(x^T)$ and taking a union bound gives
\begin{equation}\label{eq:appendix-uniformtail}
\Prb_\sigma\left\{
\max_{v\in C_T(x^T)}L_T(v;\sigma)
>6d\harm_T+\log N(x^T)+\log(1/\delta)
\right\}
\leq\delta.
\end{equation}

For the expectation bound, use log-sum-exp, Jensen's inequality, and \eqref{eq:appendix-fixedmgf}:
\begin{align}
\E_\sigma\max_{v\in C_T(x^T)}L_T(v;\sigma)
&\leq\E_\sigma\log\sum_{v\in C_T(x^T)}e^{L_T(v;\sigma)}\\
&\leq\log\sum_{v\in C_T(x^T)}\E_\sigma e^{L_T(v;\sigma)}\\
&\leq6d\harm_T+\log N(x^T).\label{eq:appendix-uniformmean}
\end{align}

Now let $X^T$ have an exchangeable law $P$, and let $\sigma$ be an independent uniform permutation. For the bad-event indicator in \eqref{eq:appendix-uniformtail}, exchangeability and Fubini's theorem give
\[
\E_X B(X^T)
=\E_X\E_\sigma B(X_{\sigma(1)},\ldots,X_{\sigma(T)})
\leq\delta.
\]
Similarly, applying \eqref{eq:appendix-uniformmean} conditionally on $X^T$ and using $N(X^T)\leq\Pi_{\Hc}(T)$ gives
\[
\E_X\sup_{h\in\Hc}L_T(h;X^T)
\leq6d\harm_T+\log\Pi_{\Hc}(T).
\]
This proves Proposition~\ref{prop:traceloss}.

\section{Arithmetic-Interval Conversion}\label{app:arithmetic}

We give the complete conversion from the trace probabilities to a finite deterministic cover.

For a fixed feature prefix, assign the empty history the interval
\[
I_\varnothing=[0,1).
\]
Suppose a realizable history $y^{t-1}$ has been assigned the interval $I_{y^{t-1}}=[a,b)$. Put
\[
c=a+q_t(0\mid x^t,y^{t-1})(b-a)
\]
and define
\[
I_{y^{t-1}0}=[a,c),
\qquad
I_{y^{t-1}1}=[c,b).
\]
A zero-probability child receives the empty interval. For a realizable trace $v^h_T$, induction gives
\[
|I_{v^h_T}|
=\prod_{t=1}^Tq_t(h(x_t)\mid x^t,v^h_{t-1})
=e^{-L_T(h;x^T)}.
\]

Let
\[
L_*=6d\harm_T+\log\Pi_{\Hc}(T)+\log(1/\delta),
\qquad
M=\lceil2e^{L_*}\rceil,
\]
and define the fixed grid
\[
U=\left\{\frac{k}{M}:k=0,\ldots,M-1\right\}.
\]
Every half-open interval $J=[a,b)\subseteq[0,1)$ of length at least $e^{-L_*}$ contains an element of $U$. Indeed, $|J|\geq2/M$. If $k=\lceil Ma\rceil$, then
\[
a\leq k/M<a+1/M<b.
\]
The case $k=M$ would imply $1-a<1/M$, contradicting $|J|\geq2/M$.

For each $u\in U$, define a deterministic sequential function $g_u$ recursively. Given $x^t$, first reconstruct
\[
g_u(x^1),\ldots,g_u(x^{t-1})
\]
and the corresponding current interval. Split that interval using \eqref{eq:q} and output the bit whose child interval contains $u$. The half-open children are disjoint and partition the parent, so the output is unique. Starting from the empty history, induction shows that the generated history remains realizable: an infeasible child has probability zero and hence an empty interval.

On the event
\[
\sup_{h\in\Hc}L_T(h;X^T)\leq L_*,
\]
every target terminal interval has length at least $e^{-L_*}$ and contains a grid point $u$. Nestedness then implies
\[
g_u(X^t)=h(X_t)
\qquad\text{for all }t\leq T.
\]
Thus $\Gc=\{g_u:u\in U\}$ is an exact stochastic global sequential cover. Finally,
\[
|\Gc|\leq M\leq2e^{L_*}+1\leq3e^{L_*},
\]
which proves \eqref{eq:coverbound}. The simpler bound \eqref{eq:coverboundsimple} follows from Sauer--Shelah and $\harm_T\leq1+\log T$.

\section{Direct Mixture Implementation}
\label{app:mixture}

We now give the direct implementation used in Theorem~\ref{thm:planar}.
Rather than enumerating the sequential cover, we maintain a mixture over
the realizable traces of the observed feature prefix.

For $v=(v_1,\ldots,v_t)\in C_t(x^t)$, define its sequential probability
under \eqref{eq:q} by
\[
\pi_t(v\mid x^t)
=
\prod_{s=1}^t
q_s(v_s\mid x^s,v_{1:s-1}).
\]
Conditional on a trace $v$, we use independent beta--Bernoulli mixtures
for the labels observed on its two sides. For $c\in\{0,1\}$, let
\[
n_c(v)=|\{s\leq t:v_s=c\}|,
\qquad
s_c(v;y^t)=\sum_{s\leq t:\,v_s=c}y_s,
\]
and define
\[
B_v(y^t)
=
\prod_{c=0}^1
\int_0^1
p^{s_c(v;y^t)}
(1-p)^{n_c(v)-s_c(v;y^t)}
\,dp.
\]
Using the beta integral,
\[
B_v(y^t)
=
\prod_{c=0}^1
\frac{1}
{(n_c(v)+1)\binom{n_c(v)}{s_c(v;y^t)}}.
\]
We combine the trace and label probabilities through
\begin{equation}
\label{eq:mixture}
Q(y^t\Vert x^t)
=
\sum_{v\in C_t(x^t)}
\pi_t(v\mid x^t)B_v(y^t).
\end{equation}

We first verify that $Q$ is a sequential probability assignment and
derive an explicit recursion. Let $v=ub\in C_t(x^t)$, where
$u\in C_{t-1}(x^{t-1})$ and $b\in\{0,1\}$. By construction,
\[
\pi_t(ub\mid x^t)
=
\pi_{t-1}(u\mid x^{t-1})
q_t(b\mid x^t,u).
\]
Moreover, conditional on the child trace $v=ub$, the beta--Bernoulli
predictive probability of $Y_t=1$ is
\begin{equation}
\label{eq:mixture-local-p}
p_t(v)
=
\frac{1+\sum_{s<t:\,u_s=b}y_s}
{2+|\{s<t:u_s=b\}|}.
\end{equation}
Consequently,
\[
B_v(y^t)
=
B_u(y^{t-1})
p_t(v)^{y_t}
(1-p_t(v))^{1-y_t}.
\]
In particular,
\begin{equation}
\label{eq:appendix-betaconsistency}
\sum_{y_t\in\{0,1\}}
B_v(y^t)
=
B_u(y^{t-1}).
\end{equation}
Summing first over the two possible outcomes and then over the feasible
children of each parent gives
\[
\sum_{y_t\in\{0,1\}}
Q(y^t\Vert x^t)
=
Q(y^{t-1}\Vert x^{t-1}).
\]
Thus $Q$ defines a sequential probability assignment, and the
corresponding prediction is
\[
\widehat p_t
=
\frac{Q(y^{t-1}1\Vert x^t)}
{Q(y^{t-1}\Vert x^{t-1})}.
\]

We next write this prediction as an explicit online recursion. After
observing $(x^{t-1},y^{t-1})$, let
\[
\rho_{t-1}(u)
=
\frac{
\pi_{t-1}(u\mid x^{t-1})B_u(y^{t-1})
}{
Q(y^{t-1}\Vert x^{t-1})
}
\]
be the posterior mass of
$u\in C_{t-1}(x^{t-1})$, with
$\rho_0(\varnothing)=1$. Once $x_t$ is observed, propagate this mass to
each feasible child $v=ub$ according to
\[
r_t(v)
=
\rho_{t-1}(u)q_t(b\mid x^t,u).
\]
Using the local prediction $p_t(v)$ from
\eqref{eq:mixture-local-p}, the learner predicts
\[
\widehat p_t
=
\sum_{v\in C_t(x^t)}
r_t(v)p_t(v).
\]
After observing $y_t$, the posterior is updated by
\[
\rho_t(v)
=
\frac{
r_t(v)p_t(v)^{y_t}(1-p_t(v))^{1-y_t}
}{
\widehat p_t^{\,y_t}(1-\widehat p_t)^{1-y_t}
}.
\]
The beta--Bernoulli predictions lie strictly between zero and one, so
all denominators are positive. The identities above show by induction
that this recursion implements exactly the mixture
\eqref{eq:mixture}.

\paragraph{Regret of the direct mixture.}
Sequential consistency implies that the cumulative log loss of the
predictor is $-\log Q(y^T\Vert x^T)$. Fix a comparator
$f(x)=a+(b-a)h(x)\in\Fc_{\Hc}$ and let
$v=h(x^T)$. Retaining only the term corresponding to $v$ in
\eqref{eq:mixture} gives
\[
Q(y^T\Vert x^T)
\geq
\pi_T(v\mid x^T)B_v(y^T).
\]
Let $n_c$ be the number of rounds with $v_t=c$ and $s_c$ the number of
ones observed on those rounds. Writing $\theta_0=a$ and $\theta_1=b$,
the elementary inequality
\[
\binom{n_c}{s_c}
\theta_c^{s_c}(1-\theta_c)^{n_c-s_c}
\leq1
\]
implies
\[
\frac{1}{(n_c+1)\binom{n_c}{s_c}}
\geq
\frac{
\theta_c^{s_c}(1-\theta_c)^{n_c-s_c}
}{
n_c+1
}.
\]
Since $(n_0+1)(n_1+1)\leq(T+1)^2$ and
$-\log\pi_T(v\mid x^T)=L_T(h;x^T)$, we obtain
\[
-\log Q(y^T\Vert x^T)
\leq
\sum_{t=1}^T\ell(f(x_t),y_t)
+
L_T(h;x^T)
+
2\log(T+1).
\]
Taking the infimum over $f\in\Fc_{\Hc}$ therefore gives, for every
$x^T$ and $y^T$,
\[
R_T(A;x^T,y^T;\Fc_{\Hc})
\leq
\sup_{h\in\Hc}L_T(h;x^T)
+
2\log(T+1).
\]
Taking the supremum over $y^T$, followed by expectation over the
features, and applying Proposition~\ref{prop:traceloss} proves
\eqref{eq:direct-mixture-bound} for every exchangeable feature law.

\section{Efficient Implementation for Planar Halfspaces}
\label{app:planar}

We now prove Theorem~\ref{thm:planar}. The implementation has two
components: enumerating all realizable labelings of the observed
feature prefix and computing their cube weights.

Let $x_i=(a_i,b_i)\in\R^2$. We first show that the trace class
$C_t(x^t)$ can be enumerated exactly. Under the convention
\[
h(x)=\one\{w^\top x+c\geq0\},
\]
every nonconstant feasible labeling of a finite sample can be realized
with a strict margin. Indeed, the negatively labeled points already
satisfy strict inequalities, while increasing $c$ by a sufficiently
small amount makes all positive inequalities strict without changing
any negative label.

A strict labeling is stable under a sufficiently small perturbation of
the normal vector. Hence it suffices to consider normals with nonzero
first coordinate. Up to scaling and complementing the labeling, such a
normal can be written as $(1,s)$. For a fixed $s$, the ordering of the
projected values
\[
a_i+s b_i
\]
changes only when two projections coincide, namely at one of the
critical values
\[
s
=
-\frac{a_i-a_j}{b_i-b_j},
\qquad
b_i\neq b_j.
\]
There are $O(t^2)$ such values. Choose one rational representative from
each interval between consecutive distinct critical values and from the
two unbounded intervals. For each representative direction, sort the
projected values and enumerate all threshold cuts and their
complements. Equal projected values are kept together, and cuts are made
only between consecutive distinct values. Since every strictly feasible labeling persists on an open
set of directions, this procedure finds every nonconstant realizable
labeling. Constant labelings are added separately, and duplicate traces
are removed. The $O(t^3)$ candidate traces can be generated and
deduplicated in $O(t^4)$ operations by storing their $t$ bits in a binary trie.

The number of distinct labelings is at most the classical affine
halfspace growth bound
\[
|C_t(x^t)|
\leq
2\sum_{k=0}^2\binom{t-1}{k}
=
t^2-t+2.
\]
Degeneracies and repeated feature points can only reduce this number.

It remains to compute the cube weights required by \eqref{eq:q}.
Planar affine halfspaces have VC dimension three, so every centered cube
contains at most three coordinates. For a fixed trace
$v\in C_t(x^t)$, it therefore suffices to enumerate the
\[
\sum_{k=0}^3\binom{t}{k}
=
O(t^3)
\]
coordinate sets of size at most three. To test the candidate cubes,
assign an integer identifier to each trace and precompute a table of
one-coordinate-flip neighbors. Comparing every pair of the $O(t^2)$
traces in $O(t)$ operations identifies every such neighbor and takes
$O(t^5)$ operations in total. A table entry records the neighboring
trace's identifier, or records that the neighbor is absent.
For each subset $A\subseteq S$ of a candidate set $S$, start at $v$
and follow the neighbor-table entries for the coordinates of $A$ in a
fixed order. This checks all $2^{|S|}\leq8$ corners using at most
$\sum_{A\subseteq S}|A|\leq12$ table accesses; a missing neighbor makes
the candidate infeasible. Thus all cube weights can be computed in
\[
O(t^2)\cdot O(t^3)
=
O(t^5)
\]
operations at round $t$. Once these weights are known, the transition
probabilities in \eqref{eq:q} and the mixture recursion of
Appendix~\ref{app:mixture} require only lower-order work. Summing over
$t\leq T$ gives
\[
\sum_{t=1}^T O(t^5)
=
O(T^6)
\]
combinatorial and rational-arithmetic operations.

Finally, suppose the input coordinates are rational. The critical
slopes are rational, and rational representatives between consecutive
critical slopes can be chosen with bit length polynomial in the input
bit length. The cube weights are integers of size $O(t^3)$, so each
transition probability in \eqref{eq:q} has numerator and denominator
of bit length $O(\log(t+1))$. The product defining $\pi_t(v\mid x^t)$
therefore has numerator and denominator of bit length
$O(t\log(t+1))$. Since $\binom ns\leq2^n$, the beta--Bernoulli factor
$B_v(y^t)$ has numerator and denominator of bit length $O(t)$.
Each unnormalized mass $\pi_t(v\mid x^t)B_v(y^t)$ consequently has
$O(t\log(t+1))$-bit components. Summing at most $O(t^2)$ such masses
gives $Q(y^t\Vert x^t)$ with numerator and denominator of bit length
$O(t^3\log(t+1))$, even using the product of the individual denominators
as a common denominator. The posterior masses and prediction probabilities
are ratios of these quantities, so their reduced fractions also have
polynomial bit length. Performing the rational arithmetic with reduction
after each operation therefore yields an exact implementation with bit
complexity polynomial in $T$ and the input bit length.

\section{A Matching Lower Bound for Exact Binary Covers}\label{app:lower}

Fix $d\geq1$ and let $\Xc=[0,1]$. Consider
\[
\Hc_d=\{x\mapsto\one\{x\in B\}:B\subseteq[0,1],\ |B|\leq d\}.
\]
This class has VC dimension $d$. Let $\mu$ be any nonatomic distribution on $[0,1]$. With probability one, $X_1,\ldots,X_T$ are distinct. On every such sample, the trace class consists of all binary vectors with at most $d$ ones and hence has cardinality
\[
\sum_{k=0}^{\min\{d,T\}}\binom Tk.
\]

On a fixed sample, one deterministic sequential function produces one binary trace. Therefore an exact cover that simultaneously covers all hypotheses must contain at least as many functions as there are distinct traces. If its success probability is positive, this lower bound holds on at least one sample in the success event, and consequently
\[
|\Gc|\geq\sum_{k=0}^{\min\{d,T\}}\binom Tk.
\]
For $T\geq2d$,
\[
\log|\Gc|\geq\log\binom Td\geq d\log(T/d).
\]
For fixed positive $d$, this matches the $\log T$ order of Theorem~\ref{thm:cover}. It does not establish optimal dependence on jointly growing $d$ and $T$.

\section{Proof of the Pseudo-Dimension Separation}
\label{app:pdimlower}

We prove Theorem~\ref{thm:pdim} in the following stronger form.

\begin{theorem}
\label{thm:appendix-pdim}
There are universal constants $c>0$ and $T_0\geq1$ such that the class
$\Fc_*$ in \eqref{eq:sigmoidclass} has pseudo-dimension one and satisfies
the following. For every $T\geq T_0$ and every
$\alpha\in(0,1/64]$, every distribution-blind uniform stochastic
sequential $\alpha$-cover $\Gc$ of $\Fc_*$ with failure probability
$1/4$ satisfies
\[
\log|\Gc|
\geq
c\log T\log(1/\alpha).
\]
\end{theorem}

The proof has three ingredients. First, although $\Fc_*$ has
pseudo-dimension one, a suitable feature can force the value of
$f_\theta$ to reveal one of $K\asymp1/\alpha$ well-separated
possibilities. Second, we arrange these $K$-way choices along a hidden
tree and choose an i.i.d.\ feature distribution supported on one hidden
root-to-leaf path. Whenever the sampled depth reaches a new record, a
sequential predictor must guess a fresh hidden digit. Third, a random
permutation has order $\log T$ records, which yields the additional
$\log T$ factor.

\subsection{Pseudo-dimension one}

Recall that
\[
\Fc_*
=
\left\{
(a,b)\mapsto \sigma(a\theta+b):
\theta\in(0,1)
\right\},
\qquad
a>0.
\]
Fix a feature $(a,b)$ and a threshold $r\in(0,1)$. Since $\sigma$ is
strictly increasing,
\[
\one\{\sigma(a\theta+b)\geq r\}
=
\one\left\{
\theta
\geq
\frac{\operatorname{logit}(r)-b}{a}
\right\}.
\]
Thus, for any finite collection of features and thresholds, the induced
binary label vector is obtained by varying a single threshold parameter
$\theta$ on the real line. These binary patterns form a chain and
cannot shatter two coordinates. Hence
\[
\pdim(\Fc_*)\leq1.
\]
Conversely, for example, the single feature $(1,0)$ with threshold
$\sigma(1/2)$ realizes both binary labels as $\theta$ varies in
$(0,1)$. Therefore
\[
\pdim(\Fc_*)=1.
\]

\subsection{Encoding a hidden \texorpdfstring{$K$}{K}-ary path}

Fix $\alpha\leq1/64$ and let
\[
K
=
\left\lfloor\frac{1}{16\alpha}\right\rfloor.
\]
Then $K\geq4$. Define $K$ closed intervals
\[
J_k
=
\left[
\frac14+4\alpha(k-1),
\frac14+4\alpha(k-1)+\alpha
\right],
\qquad
k=1,\ldots,K.
\]
They all lie in $(0,1)$, and two distinct bands are separated by more
than $2\alpha$. Consequently,
\begin{equation}
\label{eq:bands-unique}
\text{for every }z\in\R,\qquad
\operatorname{dist}(z,J_k)\leq\alpha
\text{ for at most one }k.
\end{equation}

We now recursively construct a full $K$-ary tree of nested intervals in
the parameter space $(0,1)$. Let
\[
I_\varnothing=(0,1).
\]
Suppose the interval
\[
I_u=(\ell_u,r_u)
\]
has already been associated with a node $u$. Choose numbers
\[
z_-
<
\operatorname{logit}(\min J_1),
\qquad
z_+
>
\operatorname{logit}(\max J_K),
\]
and choose $a_u>0$ and $b_u\in\R$ so that the affine map
\[
\phi_u(\theta)
=
a_u\theta+b_u
\]
maps $I_u$ bijectively onto $(z_-,z_+)$. For
$k=1,\ldots,K$, define
\[
I_{uk}
=
\phi_u^{-1}
\bigl(
\operatorname{logit}(\operatorname{int}J_k)
\bigr).
\]
The intervals $I_{u1},\ldots,I_{uK}$ are nonempty and pairwise
disjoint, and their closures are contained in $I_u$.

Associate with node $u$ the feature
\[
x_u=(a_u,b_u).
\]
By construction,
\begin{equation}
\label{eq:appendix-band}
\theta\in I_{uk}
\quad\Longrightarrow\quad
f_\theta(x_u)
=
\sigma(a_u\theta+b_u)
\in\operatorname{int}J_k.
\end{equation}

Thus, evaluating at $x_u$ reveals which one of the $K$ child intervals
contains $\theta$, up to accuracy $\alpha$.

Fix a depth $L$ and let
\[
B_1,\ldots,B_L
\stackrel{\mathrm{iid}}{\sim}
\operatorname{Unif}\{1,\ldots,K\}.
\]
The vector
\[
B=(B_1,\ldots,B_L)
\]
specifies a random root-to-leaf path. Choose an arbitrary point
\[
\theta_B
\in
I_{B_1\cdots B_L}.
\]
For $r=1,\ldots,L$, define
\[
x_r(B)
=
x_{B_1\cdots B_{r-1}}.
\]
Since the terminal interval is contained in all of its ancestors,
\eqref{eq:appendix-band} gives
\begin{equation}
\label{eq:appendix-targetband}
f_{\theta_B}(x_r(B))
\in
J_{B_r},
\qquad
r=1,\ldots,L.
\end{equation}

For each fixed path $B$, define the feature distribution
\[
\mu_B
=
\frac1L\sum_{r=1}^L\delta_{x_r(B)}.
\]
Equivalently, let
\[
D_1,\ldots,D_T
\stackrel{\mathrm{iid}}{\sim}
\operatorname{Unif}\{1,\ldots,L\}
\]
and set
\[
X_t=x_{D_t}(B).
\]
Conditional on $B$, the variables $X_1,\ldots,X_T$ are therefore
i.i.d.\ from the fixed distribution $\mu_B$.

The role of distribution-blindness is important here. The cover
$\Gc$ must be fixed before the feature distribution is chosen, so the
\emph{same} family $\Gc$ must work simultaneously for every
$\mu_B$. This allows us to randomize the hidden path $B$ after fixing
the cover.

\subsection{Every record forces a fresh \texorpdfstring{$K$}{K}-way guess}

Call round $t$ a \emph{record time} if
\[
D_t>\max_{s<t}D_s,
\]
with the first round counted as a record. Let
\[
R(D^T)
\]
denote the total number of record times. The next lemma is the core of the argument.

\begin{lemma}
\label{lem:record-fresh-digit}
Fix the depth sequence $D^T$ and a deterministic sequential rule
\[
g:\bigcup_{t=1}^T\Xc^t\to[0,1].
\]
Then
\begin{equation}
\label{eq:appendix-oneg}
\Prb_B\left\{
\max_{t\leq T}
\left|
g(X^t)-f_{\theta_B}(X_t)
\right|
\leq\alpha
\;\middle|\;
D^T
\right\}
\leq
K^{-R(D^T)}.
\end{equation}
\end{lemma}

\begin{proof}
Let
\[
t_1<\cdots<t_R
\]
be the record times and write
\[
r_j=D_{t_j}.
\]
Then
\[
r_1<r_2<\cdots<r_R.
\]

Consider a record time $t_j$ with depth $r=r_j$. Since every previously
sampled depth is strictly smaller than $r$, each feature in the prefix
\[
X^{t_j}
\]
depends only on
\[
B_1,\ldots,B_{r-1}.
\]
Indeed,
\[
x_q(B)=x_{B_1\cdots B_{q-1}}
\]
depends only on the first $q-1$ digits, and every depth observed up to
time $t_j$ is at most $r$. Therefore
\[
g(X^{t_j})
\]
is a function of $D^T$ and $B_1,\ldots,B_{r-1}$, and in particular
does not depend on the fresh digit $B_r$.

Define a coarse guess $\widehat B_r$ from $g(X^{t_j})$ as follows:
if there is a unique $k$ such that
\[
\operatorname{dist}(g(X^{t_j}),J_k)\leq\alpha,
\]
set $\widehat B_r=k$; otherwise set $\widehat B_r=0$.
Uniqueness follows from \eqref{eq:bands-unique}. The random variable
$\widehat B_r$ is measurable with respect to
\[
D^T,B_1,\ldots,B_{r-1}.
\]

If $g$ approximates the target at round $t_j$, then by
\eqref{eq:appendix-targetband},
\[
f_{\theta_B}(X_{t_j})
=
f_{\theta_B}(x_r(B))
\in J_{B_r}.
\]
Hence
\[
\left|
g(X^{t_j})-f_{\theta_B}(X_{t_j})
\right|
\leq\alpha
\quad\Longrightarrow\quad
\widehat B_r=B_r.
\]
Thus the event that $g$ approximates the target on \emph{all} rounds is
contained in
\[
\bigcap_{j=1}^R
\{\widehat B_{r_j}=B_{r_j}\}.
\]

Now reveal the digits in increasing order of the record depths. At depth
$r_j$, the guess $\widehat B_{r_j}$ is determined by digits with index
strictly smaller than $r_j$, whereas $B_{r_j}$ is still independent
and uniform on $\{1,\ldots,K\}$. Consequently,
\[
\Prb\left(
\widehat B_{r_j}=B_{r_j}
\mid
D^T,B_1,\ldots,B_{r_j-1}
\right)
\leq
\frac1K.
\]
Applying the chain rule over the $R$ record depths gives
\[
\Prb_B\left(
\bigcap_{j=1}^R
\{\widehat B_{r_j}=B_{r_j}\}
\;\middle|\;
D^T
\right)
\leq
K^{-R}.
\]
This proves \eqref{eq:appendix-oneg}.
\end{proof}

The proof deliberately works with the coarse events
$\{\widehat B_r=B_r\}$ rather than conditioning on the exact
approximation events themselves. An exact approximation event can
depend on the precise value of $f_{\theta_B}$ and hence on deeper path
digits, whereas the coarse guess at a record depth depends only on the
already exposed prefix.

\subsection{From one rule to a cover}

Let
\[
M=|\Gc|.
\]
Fix $D^T$. By Lemma~\ref{lem:record-fresh-digit} and a union bound over
the $M$ sequential rules,
\begin{equation}
\Prb_B\left\{
\exists g\in\Gc:
\max_{t\leq T}
|g(X^t)-f_{\theta_B}(X_t)|
\leq\alpha
\;\middle|\;
D^T
\right\}
\leq
\min\{1,MK^{-R(D^T)}\}.
\label{eq:appendix-uniong}
\end{equation}

We now use the defining quantifiers of a distribution-blind cover.
For every fixed path $B$, the same family $\Gc$ must be an
$\alpha$-cover under the distribution $\mu_B$. Therefore, for the
particular target $f_{\theta_B}$,
\[
\Prb_{D^T}\left\{
\exists g\in\Gc:
\max_{t\leq T}
|g(X^t)-f_{\theta_B}(X_t)|
\leq\alpha
\right\}
\geq
\frac34.
\]
Averaging this inequality over the random path $B$ and exchanging the
order of expectation gives
\begin{equation}
\label{eq:appendix-coveraverage}
\frac34
\leq
\E_{D^T}
\min\{1,MK^{-R(D^T)}\}.
\end{equation}

Thus all that remains is to understand the number of records in an
i.i.d.\ sequence of random depths.

\subsection{The record count}

Let
\[
\mathcal C_L
=
\{D_1,\ldots,D_T\text{ are all distinct}\}.
\]
A union bound gives
\[
\Prb(\mathcal C_L^c)
\leq
\frac{T(T-1)}{2L}.
\]
Conditional on $\mathcal C_L$, the relative ordering of
$D_1,\ldots,D_T$ is a uniformly random permutation of $[T]$.
The record count depends only on this relative ordering. Hence, if
$R_T$ denotes the number of records in a uniformly random permutation
of $[T]$, then for every function $\psi:\{1,\ldots,T\}\to[0,1]$,
\[
\left|
\E\psi(R(D^T))
-
\E\psi(R_T)
\right|
\leq
\Prb(\mathcal C_L^c).
\]
Since \eqref{eq:appendix-coveraverage} holds for every $L$, letting
$L\to\infty$ with
\[
\psi(r)=\min\{1,MK^{-r}\}
\]
gives
\begin{equation}
\label{eq:appendix-permutationrecords}
\frac34
\leq
\E\min\{1,MK^{-R_T}\}.
\end{equation}

For completeness, recall the probability generating function of
$R_T$. If $\rho_t$ is the rank of the $t$th element among the first
$t$ elements of a uniformly random permutation, then
\[
\rho_1,\ldots,\rho_T
\]
are independent and $\rho_t$ is uniform on $\{1,\ldots,t\}$. A record
occurs exactly when $\rho_t=t$. Therefore
\begin{equation}
\label{eq:appendix-recordpgf}
\E z^{R_T}
=
\prod_{t=1}^T
\left(
1-\frac1t+\frac zt
\right).
\end{equation}

Set
\[
\lambda
=
\frac{\log2}{\log K}.
\]
Since $K\geq4$, we have $0<\lambda\leq1$. For every $u\geq0$,
\[
\min\{1,u\}
\leq
u^\lambda.
\]
Using $K^{-\lambda}=1/2$ in
\eqref{eq:appendix-permutationrecords}, followed by
\eqref{eq:appendix-recordpgf}, yields
\[
\frac34
\leq M^\lambda\E K^{-\lambda R_T}
=M^\lambda\E 2^{-R_T}
=M^\lambda\prod_{t=1}^T\left(1-\frac1{2t}\right)
\leq M^\lambda e^{-\harm_T/2}.
\]
Hence
\begin{equation}
\label{eq:appendix-pdimfinal}
\log M
\geq
\frac{\log K}{\log2}
\left(
\frac{\harm_T}{2}
+
\log\frac34
\right).
\end{equation}

For $T\geq2$,
\[
\frac{\harm_T}{2}+\log\frac34
\geq
\frac{\harm_T}{4},
\]
and $\harm_T\geq\log T$. Moreover, since
$\alpha\leq1/64$,
\[
K
=
\left\lfloor\frac1{16\alpha}\right\rfloor
\geq
\frac1{32\alpha}.
\]
Because $\log(1/\alpha)\geq\log64$,
\[
\log K
\geq
\log\frac1\alpha-\log32
\geq
\frac16\log\frac1\alpha.
\]
Substituting these two bounds into
\eqref{eq:appendix-pdimfinal} gives
\[
\log M
\geq
\frac{1}{24\log2}
\log T\log(1/\alpha)
\]
for all $T\geq2$. This proves
Theorem~\ref{thm:appendix-pdim}, with a universal constant $c$.
Taking $\alpha=1/T$ proves Theorem~\ref{thm:pdim}.

\section{Measurability and Standard Examples}\label{app:measurability}
For a fixed horizon $t$, there are finitely many binary vectors $v\in\{0,1\}^t$. Assumption~\eqref{eq:measurable} says that the indicator of the event $v\in C_t(x^t)$ is measurable for each such vector. Therefore the finite-valued mapping $x^t\mapsto C_t(x^t)$ is measurable when the set of trace classes carries its discrete sigma-algebra. Cube counts, feasibility decisions, and the ratios in \eqref{eq:q} are functions of this finite object and the finite label history. Thus $q_t$ is measurable.

For a fixed seed $u$, the endpoints of its arithmetic interval are measurable functions of the feature prefix, by induction on $t$. Comparing $u$ to the split point gives a measurable bit, so $g_u$ is measurable. To see that the uniform log-loss event is measurable without taking an uncountable union, enumerate all $2^T$ binary vectors. For each $v$, its realizability and its path probability are measurable. The event is the finite union over $v$ of realizability together with the corresponding loss exceeding the threshold. The cover failure event can similarly be written as a finite union over binary traces, followed by a finite intersection over covering predictors. Supremum regret over binary label sequences is also a finite maximum for the constructed strategies.

If $\Hc$ is countable and each hypothesis is measurable, \eqref{eq:measurable} is a countable union of finite intersections of measurable sets. This is a sufficient condition, not a restriction to countable hypothesis classes. For the sparse-set class in Appendix~\ref{app:lower}, a labeling is feasible exactly when equal feature values receive equal labels and at most $d$ distinct feature values receive label one. These conditions are Borel on $[0,1]^t$.

For affine halfspaces on $\mathbb R^m$, every feasible labeling of a finite sample can also be realized with rational coefficients. To prove this, start from a realizing pair $(w,a)$ for the rule $\one\{\langle w,x\rangle\geq a\}$. Every negatively labeled sample point has a strictly negative value of $\langle w,x_i\rangle-a$. Decrease $a$ by a sufficiently small positive amount. All positive points now have a strictly positive margin, and all negative points retain a strictly negative margin. There are finitely many inequalities, so a sufficiently close rational pair preserves every sign. Thus each finite-feasibility set is a countable union over rational coefficients of Borel sets.

For completeness, affine halfspaces in $\mathbb R^m$ have VC dimension $m+1$. An affinely independent set of $m+1$ points can receive any prescribed $\{-1,1\}$ values from an affine function, and is therefore shattered. Any $m+2$ points have a nonzero affine dependence $\sum_i\lambda_i x_i=0$, $\sum_i\lambda_i=0$. There is at least one positive and one negative coefficient. Label the points with positive coefficients one and those with negative coefficients zero. A realizing affine function $f$ would satisfy $\lambda_i f(x_i)\geq0$ for positive coefficients and $\lambda_i f(x_i)>0$ for negative coefficients. This contradicts $\sum_i\lambda_i f(x_i)=0$. Coefficients equal to zero do not affect the argument.

\section{Detailed Related Work and Comparison of Protocols}\label{app:literature}

\subsection{Unknown stochastic features and adversarial labels}
The hybrid learning framework dates to \citet{lazaric2012}, whose conference precursor appeared in 2009. Their learners use empirical covers with stochastic inputs and adversarial outputs. \citet{rakhlin2011stochastic} place stochastic and constrained adversaries in a general minimax framework. Distribution knowledge matters: in our game, one sequential strategy is fixed before the choice of an arbitrary feature law. Moreover, the supremum over the entire label sequence is inside the feature expectation. Results for jointly i.i.d. labeled samples or for a known reference measure do not directly imply this guarantee.

The two-level comparator $\mathcal F_{\Hc}$ is established prior work. \citet{bhatt2021} study precisely this model under conditionally Bernoulli labels and ask for a constructive logarithmic bound for general finite-VC partitions. \citet[Theorem 11]{bhatt2023portfolio} analyze statewise categorical probability mixtures with adversarial outcomes; their binary two-state case is especially close to our application. Their general i.i.d. bound retains an empirical-process term of square-root order. Thus neither the partition model nor the use of independent Bernoulli mixtures is introduced here. The new ingredient is the cube-weighted sequential distribution over latent traces and its simultaneous random-order guarantee.

\citet[Example 2, Proposition 2, and Corollary 2]{wu2023} give the direct stochastic-cover predecessor for this same expected-worst-case log-loss problem. Their general VC rate is $O((d+1)\log^2 T)$. They already obtain logarithmic covers for subclasses controlled by star-type dimensions and for certain interval and cuboid compositions. Our binary result applies without these additional structural conditions. The finite-VC polynomial-cover question in their Remark 4, including planar halfspaces, also appears in \citet[Remark 7.1]{wu2026universal}.

\citet{bhatt2023smoothed} obtain logarithmic regret for two-level VC models under smoothed features relative to a \emph{known} reference distribution; at smoothing parameter one, this includes known-law i.i.d. features. Their transductive reduction does not provide the single distribution-blind strategy required here. Extensions of stochastic covering to unknown changing feature distributions appear in \citet{wu2023dynamic}. Further applications include private-label functional distributions \citep{wu2023private} and hybrid-learning reductions for omniprediction \citep{okoroafor2025omni}. These results establish the broader relevance of the cover notion, while using different loss or observation models.

\subsection{One-inclusion geometry, compression, and computation}
The random-order viewpoint originates in one-inclusion prediction \citep{haussler1994}. The trace growth estimates are the classical Sauer--Shelah inequalities \citep{sauer1972,shelah1972}. Cubical structure, shifting, and compression have been studied extensively \citep{rubinstein2009,chalopin2019}. In that language, $W_C(v)$ counts all cubes of the trace class incident to $v$, including its zero-dimensional cube. These objects are classical; the contribution here is their use as probability weights and the pointwise coordinate-deletion inequality of Lemma~\ref{lem:deletion}. It does not require the trace class to be maximum, ample, or to admit a corner peeling. Recent agnostic PAC work also uses Boolean-cube orientations and leave-one-out analysis \citep{mathiasen2026}; its target is classification risk under jointly i.i.d. data rather than our uniform cumulative log-loss bound.

The multiclass one-inclusion literature deserves particular care. \citet{rubinstein2009} bound one-inclusion hypergraph density by Pollard pseudo-dimension. \citet{simon2010} repair the proof, showing that the original assertion that shifting preserves pseudo-dimension is false. The corrected density result concerns expected categorical prediction errors. It does not by itself bound the uniform pathwise probability cost needed by our covering argument. Our lower bound is consistent with that result: a fresh probability value can require choosing among many separated bands.

Arithmetic intervals \citep{witten1987} and mistake-location expert reductions \citep{bendavid2009} are established coding devices. Our conversion uses a common fixed seed grid with the cube predictor's uniform log-loss bound. Similarly, planar trace counts and arrangement enumeration are classical \citep{cover1965,edelsbrunner1986}. The computational contribution is their combination with exact cube weights and the posterior recursion, including polynomial bit complexity for rational planar features. The method is polynomial for fixed dimension; its degree grows with dimension. Oracle-efficient unknown-law hybrid learning \citep{wu2024} and the more recent constrained-label setting of \citet{okoroafor2026hybrid} address a different computational model. Simulator-based classification \citep{voitovych2026simulators} also gives the learner additional access to the feature process. These results do not justify a lower bound on the running time of earlier log-loss algorithms, and we make no such claim.

\subsection{Pseudo-dimension, covering entropy, and log-loss}
Classical scale-sensitive dimensions control empirical covers and learnability for real-valued classes \citep{alon1997}. Gapped variants and their relation to empirical and sequential entropy are studied by \citet{jia2025}. The distinctions between a cover constructed after a fixed design is known, a cover on a specified feature tree \citep{rakhlin2015sequential}, and one fixed family of sequential rules valid for every unknown i.i.d. law are essential. The lower bound in Theorem~\ref{thm:pdim} concerns the last notion. It therefore does not contradict small ordinary empirical covers for finite pseudo-dimension classes.

Random-branch lower bounds in distribution-dependent learning use an unseen deeper tree node to force a fresh prediction \citep{bousquet2021universal,bousquet2023curves}. Our construction uses this general principle with nested parameter intervals realizing $K$ separated sigmoid bands. It counts repeated record depths and bounds the success of an entire finite sequential cover. This produces the additional $\log K$ cost within one fixed, ordered sigmoid family. The feature distribution may depend on the horizon and chosen branch; the function class does not. We do not assert a single hard distribution simultaneously for all horizons.

Covering entropy provides useful upper bounds for log-loss but need not characterize its exact minimax regret. Sequential probability assignment and refined entropy bounds are developed by \citet{rakhlin2015probability,bilodeau2020}; smooth truncation and Bayesian covers are analyzed by \citet{wu2023bayesian}. More recent square-root entropy bounds are tailored to the geometry of probability assignment \citep{jia2025entropy}. These results motivate retaining the distinction between our uniform approximation lower bound and a binary log-loss lower bound. We have not transferred any adversarial-tree entropy characterization to the distribution-blind i.i.d. game without proof. Bayesian aggregation and mixability themselves are classical \citep{vovk1990,cesabianchi2006}.

\subsection{Logistic regret and efficient improper learning}
Logarithmic logistic regret from Bayesian prediction predates the present work \citep{kakade2004}. Lower bounds for proper bounded-domain learning \citep{hazan2014} motivate the improper strategies of \citet{foster2018}. The computational line includes binary logistic prediction \citep{jezequel2020}, efficient multiclass aggregation \citep{jezequel2021,agarwal2022}, and the recent mixture-of-sigmoids implementation \citep{digennaro2026}. These works provide increasingly efficient algorithms under norm-dependent guarantees. \citet{shamir2020} analyze how the norm constraint and dimension affect regret. Consequently our planar theorem should be read as a guarantee for two-level partitions; it does not replace these algorithms for sigmoid probabilities.

\citet{mayo2022} adapt to unknown comparator and feature scales. Their scale-free bounds still depend on the realized scale, including the comparator--feature norm product. The absence of an advance norm bound is therefore distinct from a regret guarantee depending only on dimension and horizon. A dimension-only logarithmic bound for unrestricted logistic regression under unknown, causally revealed feature laws remains open.

\subsection{Full-design logistic regret and statistical risk}
In the transductive model, the entire feature multiset is available before prediction. \citet[Theorem 13]{qian2026} obtain unrestricted logistic regret at most $6q\log(eT)$ in this model, with adversarial labels and no feature or parameter norm constraint. Their strategy can be approximated in polynomial time at fixed dimension using slab enumeration and log-concave sampling. It is an especially important comparison to both our sequential planar implementation and our unresolved unrestricted-logit question: the missing assumption here is advance access to the feature multiset, rather than simply computational power.

Precise logistic regret through Shtarkov sums has a substantial fixed-design literature, including lower bounds \citep{shamir2021lower}, categorical and general feature asymptotics \citep{jacquet2021,jacquet2022}, unbounded weights \citep{drmota2024}, regularized unbounded weights \citep{drmota2025}, and large-weight phase transitions \citep{drmota2026}. The latter also averages fixed-design regret over certain random designs; that ordering still gives the learner the full design in each game. The recent preprint of \citet[Theorem 4.1]{chardon2026} bounds the unrestricted logistic Shtarkov sum for arbitrary fixed designs by $\sum_{j=0}^q\binom{T}{j}$, where $q$ is the parameter dimension. Such design-dependent probability assignments do not directly give a sequential strategy for unknown i.i.d. features.

Statistical excess risk is a further distinct target. \citet{bilodeau2023density} analyze well-specified conditional density estimation through empirical entropy. \citet{mourtada2022smp} give an improper estimator for misspecified density estimation and logistic regression, and \citet{qian2026} derive norm-independent batch risk bounds through transductive priors. Jointly i.i.d. labeled data permit statistical reductions unavailable when the labels are chosen adversarially after inspecting the features. These batch results therefore concern a distinct protocol. In addition, the efficient specialization of \citet{qian2026} assumes a well-specified logistic response model and regular covariate distributions.

\end{document}